\documentclass{article}
\usepackage{times}
\usepackage{iclr2024_conference}[camera-ready]
\iclrfinalcopy
\usepackage[utf8]{inputenc} % allow\ utf-8 input
\usepackage[T1]{fontenc}    % use 8-bit T1 fonts
\usepackage[pagebackref=true]{hyperref}       % hyperlinks
\usepackage{url}            % simple URL typesetting
\usepackage{booktabs}       % professional-quality tables
\usepackage{amsfonts}       % blackboard math symbols
\usepackage{nicefrac}       % compact symbols for 1/2, etc.
\usepackage{microtype}      % microtypography
\usepackage{xcolor}         % colors
\usepackage{xspace}
\usepackage{soul}
\usepackage{caption}
\usepackage{subcaption,siunitx,booktabs}
\usepackage{array}

\usepackage{mathtools}
\usepackage{bm}
\usepackage[capitalize,noabbrev, nameinlink]{cleveref}
\usepackage{graphicx}
\usepackage{algorithm}
\usepackage{algpseudocode}
\usepackage{enumitem}
\usepackage{wrapfig}
\usepackage{pifont}
\usepackage{multirow}
\usepackage{placeins}
\usepackage{float} 

\usepackage{amsmath,amsfonts,bm}

\def\eqref#1{equation~\ref{#1}}
\def\1{\bm{1}}

\DeclareMathAlphabet{\mathsfit}{\encodingdefault}{\sfdefault}{m}{sl}
\SetMathAlphabet{\mathsfit}{bold}{\encodingdefault}{\sfdefault}{bx}{n}

\usepackage{url}
\usepackage[normalem]{ulem}

\definecolor{BrickRed}{rgb}{0.6,0,0}
\definecolor{RoyalBlue}{rgb}{0,0,0.8}
\definecolor{Tdgreen}{rgb}{0,0.4,0.7}
\definecolor{cadmiumgreen}{rgb}{0.0, 0.42, 0.24}
\hypersetup{colorlinks, citecolor=RoyalBlue, linkcolor=RoyalBlue}

\newcommand{\cmark}{\textcolor{green!80!black}{\ding{51}}}
\newcommand{\xmark}{\textcolor{red}{\ding{55}}}

\definecolor{byzantium}{rgb}{0.44, 0.16, 0.39}
\definecolor{grn}{rgb}{0.1, 0.6, 0.1}
\definecolor{mgt}{rgb}{0.7, 0.3, 0.7}
\definecolor{chamoisee}{rgb}{0.63, 0.47, 0.35}
\definecolor{purp}{rgb}{0.65, 0.16, 0.65}
\definecolor{alizarin}{rgb}{0.82, 0.1, 0.26}
\definecolor{azure(colorwheel)}{rgb}{0.0, 0.5, 1.0}
\definecolor{brown}{rgb}{0.65, 0.16, 0.16}
\definecolor{lblue}{rgb}{0, 0.2, 0.8}
\definecolor{orange}{rgb}{1.0, 0.5, 0.0}

\definecolor{myblue}{RGB}{0, 0, 255} % main highlighting color
\definecolor{mydarkblue}{RGB}{0, 0, 139} % darker hover color
\hypersetup{
    colorlinks=true,
    linkcolor=mydarkblue,
    filecolor=mydarkblue,
    citecolor=mydarkblue,
    urlcolor=mydarkblue,
}

\def\tree{$K^{2}$--tree\xspace}
\def\trees{$K^{2}$--trees\xspace}

\def\Algname{HGGT\xspace}

\algnewcommand\algorithmicforeach{\textbf{for each}}
\algdef{S}[FOR]{ForEach}[1]{\algorithmicforeach\ #1\ \algorithmicdo}

\newcolumntype{C}[1]{>{\centering\arraybackslash}p{#1}}

\title{Graph Generation with \trees}

\author{Yunhui Jang, Dongwoo Kim, Sungsoo Ahn\\
Pohang University of Science and Technology\\
\texttt{\{uni5510, dongwookim, sungsoo.ahn\}@postech.ac.kr} \\
}

\begin{document}

\maketitle

\begin{abstract}
Generating graphs from a target distribution is a significant challenge across many domains, including drug discovery and social network analysis. In this work, we introduce a novel graph generation method leveraging \tree representation, originally designed for lossless graph compression. The \tree representation {encompasses inherent hierarchy while enabling compact graph generation}. In addition, we make contributions by (1) presenting a sequential \tree representation that incorporates pruning, flattening, and tokenization processes and (2) introducing a Transformer-based architecture designed to generate the sequence by incorporating a specialized tree positional encoding scheme. Finally, we extensively evaluate our algorithm on four general and two molecular graph datasets to confirm its superiority for graph generation.
\end{abstract}

\section{Introduction}
Generating graph-structured data is a challenging problem in numerous fields, such as molecular design \citep{li2018learning, maziarka2020mol}, social network analysis \citep{grover2019graphite}, and public health \citep{yu2020reverse}. Recently, deep generative models have demonstrated significant potential in addressing this challenge \citep{simonovsky2018graphvae, jo2022gdss, vignac2022digress}. In contrast to the classic random graph models \citep{albert2002statistical, erdHos1960evolution}, these methods leverage powerful deep generative paradigms, e.g., variational autoencoders \citep{simonovsky2018graphvae}, normalizing flows \citep{madhawa2019graphnvp}, and diffusion models \citep{jo2022gdss}.

\begin{figure*}[t]
\vspace{-0.1in}
\begin{minipage}{0.55\linewidth}
    \centering
    \includegraphics[width=0.9\linewidth]{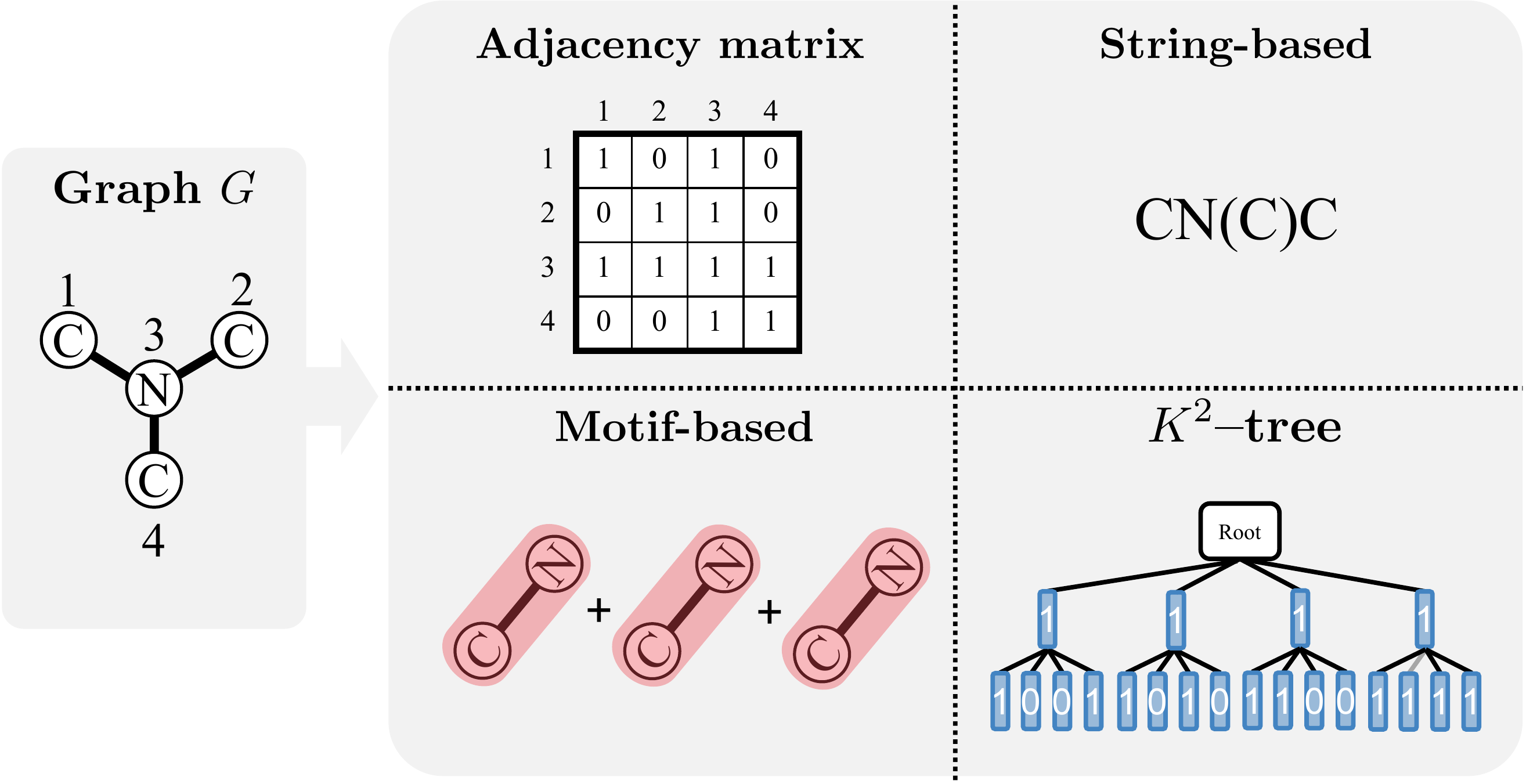}
    
\end{minipage}
% \hfill
\hspace*{\fill}
\begin{minipage}{0.45\linewidth}
    % \vspace{0.1in}
    \scalebox{0.95}{
    % \begin{table}
%   \caption{Comparison of sequential \tree representation to existing graph representations. Note that $c$ denotes the number of motifs.}
%   \label{tab:various_rep}
%   \centering
%   \scalebox{0.7}{
%     \begin{tabular}{lcccc}
%     \toprule
%     Method & Domain-agnostic & Hierarchical & Node / edge features & Complexity \\
%     \midrule
%     Adjacency matrix & \cmark & \xmark & \xmark & $O(n^2)$  \\
%     String-based & \xmark & \xmark & \cmark & $O(n+m)$ \\
%     Motif-based & \xmark & \cmark & \cmark & $O(c)$ \\
%     \tree & \cmark & \cmark & \cmark & $Km(\log_{k^2}({n^2}/{m})+O(1))$ \\
%     \textbf{Ours (sequential \tree representation)} & \cmark & \cmark & \cmark & $m(\log_{k^2}({n^2}/{m})+O(1))$ \\
%     \bottomrule
%   \end{tabular}
%   }
% \end{table}

% \begin{wraptable}{r}{0.6\textwidth}
%\begin{table}
  % \caption{Comparison of graph representations according to being domain-agnostic (DA), hierarchical (H), and able to handle attributed graphs (F).} %Note that $c$ denotes the number of motifs.}
  % \label{tab:various_rep}
  % \centering
  % \centering
  %   \begin{tabular}{lcccc}
  %   \toprule
  %   Method & DA & H & F \\ %& Complexity \\
  %   \midrule
  %   Adjacency matrix & \cmark & \xmark & \xmark \\%& $O(n^2)$  \\
  %   String-based & \xmark & \xmark & \cmark \\%& $O(n+m)$ \\
  %   Motif-based & \xmark & \cmark & \cmark \\%& $O(c)$ \\
  %   \tree & \cmark & \cmark & \cmark \\%& $Km(\log_{K^2}({n^2}/{m})+O(1))$ \\
  %   %\textbf{Ours} & \cmark & \cmark & \cmark & $m(\log_{K^2}({n^2}/{m})+O(1))$ \\
  %   \bottomrule
  % \end{tabular}
%\end{table}
% \end{wraptable}
\centering
\begin{tabular}{lccccc}
    \toprule
    Method & Repr. & H & A & DA \\
    \midrule
    GraphRNN  & Adj. & \xmark & \xmark & \cmark \\%& $O(n^2)$  \\
    GraphGen  & String & \xmark & \cmark & \cmark \\
    JT-VAE & Motif & \cmark & \cmark & \xmark \\
    GDSS & Adj. & \xmark & \cmark & \cmark \\
    \Algname (ours) & \tree & \cmark & \cmark & \cmark \\
    \bottomrule
\end{tabular}}
    \label{tab:concept}
\end{minipage}
    % \vspace{-0.15in}
    \caption{\label{fig:concept}\textbf{(Left) Various representations used for graph generation.} \textbf{(Right) Comparing graph generative methods in terms of used graph representation.} The comparison is made with respect to a method being hierarchical (H), able to handle attributed graphs (A), and domain-agnostic (DA).}
    % \vspace{1in}
    \vspace{-0.2in}
\end{figure*}

The graph generative models can be categorized into three types by the graph representation the models generate.
First, an adjacency matrix is the most common representation \citep{simonovsky2018graphvae, madhawa2019graphnvp, liu2021graphebm}. Secondly, a string-based representation extracted from depth-first tree traversal on a graph can represent the graph as a sequence \citep{ahn2022spanning, goyal2020graphgen, krenn2019selfies}. Finally, representing a graph as a composition of connected motifs,  i.e., frequently appearing subgraphs, can preserve the high-level structural properties \citep{jin2018junction, jin2020hierarchical}. We describe the representations on the left of \cref{fig:concept}. 

Although there is no consensus on the best graph representation, two factors drive their development. First is the need for compactness to reduce the complexity of graph generation {and simplify the search space over graphs}. For example, to generate a graph with $N$ vertices and $M$ edges, the adjacency matrix requires specifying $N^{2}$ elements. In contrast, the string representation typically requires specifying $O(N+M)$ elements, leveraging the graph sparsity \citep{ahn2022spanning, goyal2020graphgen, segler2018generating}. Motif representations also save space by representing frequently appearing subgraphs by basic building blocks \citep{jin2018junction, jin2020hierarchical}.

The second factor driving the development of new graph representations is the presence of a hierarchy in graphs. For instance, community graphs possess underlying clusters, molecular graphs consist of distinct chemical fragments, and grid graphs exhibit a repetitive coarse-graining structure. In this context, motif representations \citep{jin2018junction, jin2020hierarchical} address the presence of a hierarchy in graphs; however, they are limited to a fixed vocabulary of motifs observed in the dataset or a specific domain.

\textbf{Contribution.} In this paper, we propose a novel graph generation framework, coined \textbf{H}ierarchical \textbf{G}raph \textbf{G}eneration with $K^{2}$--\textbf{T}ree (\Algname), which can represent not only non-attributed graphs but also attributed graphs in a compact and hierarchical way without domain-specific rules. The right-side table of \cref{fig:concept} emphasizes the benefits of \Algname. Since the \tree recursively redefines a graph into $K^{2}$ substructures, our representation becomes more compact and enables consideration of hierarchical structure \textcolor{black}{in adjacency matrices}.\footnote{\color{black}This differs from the conventional hierarchical community structure. We provide the discussion in \cref{appx: discussion}.}

Specifically, we model the process of graph generation as an autoregressive construction of the \tree. To this end, we design a sequential \tree representation that recovers the original \tree when combined sequentially. In particular, we propose a two-stage procedure where (1) we prune the \tree to remove redundancy arising from the symmetric adjacency matrix for undirected graphs and (2) subsequently flatten and tokenize the \tree into a sequence to minimize the number of decisions required for the graph generation.

We employ the Transformer architecture \citep{vaswani2017attention} to generate the sequential \tree representation of a graph. To better incorporate the positional information of each node in a tree, we design a new positional encoding scheme specialized to the \tree structure. Specifically, we represent the positional information of a node by its pathway from the root node; the proposed encoding enables the reconstruction of the full \tree given just the positional information.

To validate the effectiveness of our algorithm, we test our method on popular graph generation benchmarks across six graph datasets: Community, Enzymes \citep{schomburg2004brenda}, Grid, Planar, ZINC \citep{irwin2012zinc}, and QM9 \citep{ramakrishnan2014quantum}. Our empirical results confirm that \Algname significantly outperformed existing graph generation methods on five out of six benchmarks, verifying the capability of our approach for high-quality graph generation across diverse applications.

To summarize, our key contributions are as follows:
\begin{itemize}[leftmargin=0.3in]
    \item We propose a new graph generative model based on adopting the \tree as a compact, hierarchical, and domain-agnostic representation of graphs.
    \item We introduce a novel, compact sequential \tree representation obtained from pruning, flattening, and tokenizing the \tree.
    \item We propose an autoregressive model to generate the sequential \tree representation using Transformer architecture with a specialized positional encoding scheme.
    \item We validate the efficacy of our framework by demonstrating state-of-the-art graph generation performance on five out of six graph generation benchmarks.
\end{itemize}
\section{Related Work}\label{sec:related}

\begin{figure}
\centering
    \includegraphics[width=0.9\textwidth]{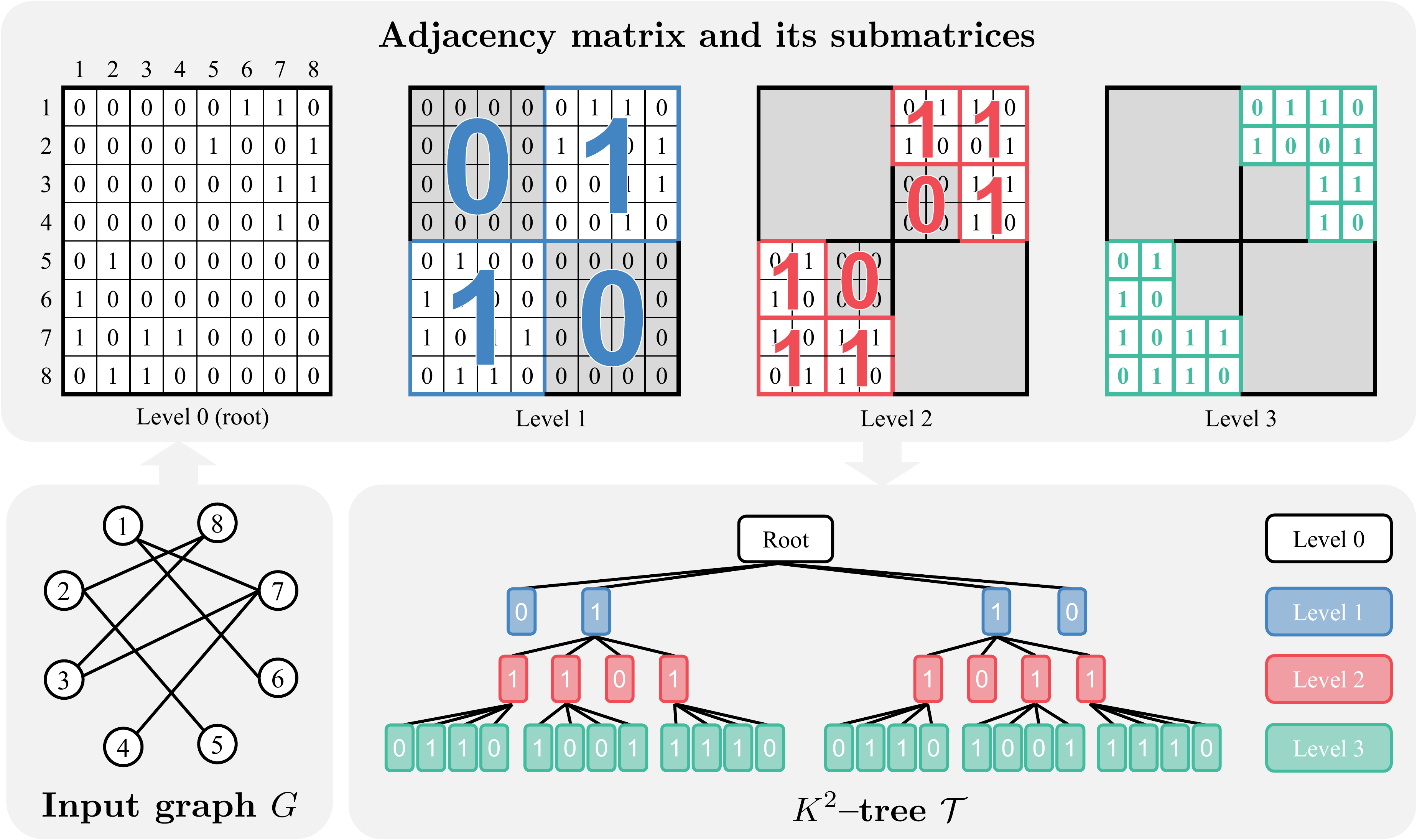}
    \caption{\textbf{\tree with $K=2$.} The \tree describes the hierarchy of the adjacency matrix iteratively being partitioned to $K\times K$ submatrices. It is compact due to summarizing any zero-filled submatrix with a size larger than $1\times 1$ (shaded in grey) by a leaf node $u$ with label $x_{u}=0$.}
    \label{fig:k2_tree}
    \vspace{-.1in}
\end{figure}

\textbf{Graph representations for graph generation.} The choice of graph representation is a crucial aspect of graph generation, as it significantly impacts the efficiency and allows faithful learning of the generative model. The most widely used one is the adjacency matrix, which simply encodes the pairwise relationship between nodes \textcolor{black}{\citep{jo2022gdss, vignac2022digress, you2018graphrnn, liao2019efficient, shi2020graphaf, luo2021graphdf, kong2023autoregressive, chen2023efficient}}. However, several methods \citep{vignac2022digress, you2018graphrnn, jo2022gdss} suffer from the high complexity in generating the adjacency matrix, especially for large graphs.

To address this issue, researchers have developed graph generative models that employ alternative graph representations such as motif-based representations and string-based representations. For instance, 
\cite{ahn2022spanning, segler2018generating} proposed to generate molecule-specific string representations, and \cite{jin2018junction, jin2020hierarchical, yang2021hit} suggested generative models that extract reasonable fragments from data and generate the set of motifs. However, these methods rely on domain-specific knowledge and are restricted to molecular data. 

\textbf{Lossless graph compression. }
Lossless graph compression \citep{besta2018survey} aims to reduce the size and complexity of graphs while preserving their underlying structures. Specifically, several works \citep{brisaboa2009k2, raghavan2003representing} introduced hierarchical graph compression methods that compress graphs leveraging their hierarchical structure. In addition, \cite{bouritsas2021partition} derived the compressed representation using a learning-based objective.

\section{\tree Representation of a Graph}\label{sec:prelim}

In this section, we introduce the \tree as a hierarchical and compact representation of graphs, as originally proposed for graph compression \citep{brisaboa2009k2}. In essence, the \tree is a $K^{2}$-ary ordered tree that recursively partitions the adjacency matrix into $K\times K$ submatrices.\footnote{By default, we assume the number of nodes in the original graph to be the power of $K^{2}$.} Its key idea is to summarize the submatrices filled only with zeros with a single tree-node, exploiting the sparsity of the adjacency matrix. From now on, we indicate the tree-node as a node. The representation is hierarchical, as it associates each parent and child node pair with a matrix and its corresponding submatrix, respectively, as described in \cref{fig:k2_tree}. 

To be specific, we consider the \tree representation $(\mathcal{T}, \mathcal{X})$ of an adjacency matrix $A$ as a $K^{2}$-ary tree $\mathcal{T} = (\mathcal{V}, \mathcal{E})$ associated with binary node attributes $\mathcal{X} = \{x_{u}: u\in\mathcal{V}\}$. Every non-root node is uniquely indexed as $(i,j)$-th child of its parent node for some $i, j \in \{1,\ldots, K\}$. The tree $\mathcal{T}$ is ordered so that every $(i, j)$-th child node is ranked $K(i-1)+j$ among its siblings. Then the \tree satisfies the following conditions:
\begin{itemize}
    \item Each node $u$ is associated with a submatrix $A^{(u)}$ of the adjacency matrix $A$.
    \item If the submatrix $A^{(u)}$ for a node $u$ is filled only with zeros, $x_{u}=0$. Otherwise, $x_{u}=1$.
    \item A node $u$ is a leaf node if and only if $x_{u}=0$ or the matrix $A^{(u)}$ is a $1\times 1$ matrix.
    \item Let $B_{1,1},\ldots, B_{K, K}$ denote the $K\times K$ partitioning of the matrix $A^{(u)}$ with $i, j$ corresponding to row- and column-wise order, respectively. The child nodes $v_{1,1},\ldots, v_{K,K}$ of the tree-node $u$ are associated with the submatrices $B_{1,1},\ldots, B_{K, K}$, respectively.
\end{itemize}
The generated \tree is a compact description of graph $G$ as any node $u$ with $x_{u}=0$ and $d_{u}<\max_u{d_u}$ where $d_u$ is the distance from the root. summarizes a large submatrix filled only with zeros. In the worst-case scenario, the size of the \tree is $MK^2(\log_{K^2}({N^2}/{M})+O(1))$ \citep{brisaboa2009k2}, where $N$ and $M$ denote the number of nodes and edges in the original graph, respectively. This constitutes a significant improvement over the $N^2$ size of the full adjacency matrix.

Additionally, the \tree is hierarchical ensuring that (1) each tree node represents the connectivity between a specific set of nodes, and (2) nodes closer to the root correspond to a larger set of nodes. We emphasize that the nodes associated with submatrices overlapping with the diagonal of the original adjacency matrix indicate intra-connectivity within a group of nodes. In contrast, the remaining nodes describe the interconnectivity between two distinct sets of nodes.

We also describe the detailed algorithms for constructing a \tree from a given graph $G$ and recovering a graph from the \tree in \cref{appx:graphtok2,appx:k2tograph}, respectively. It is crucial to note that the ordering of the nodes in the adjacency matrix influences the \tree structure. \textcolor{black}{Inspired by \cite{diamant2023improving}, we adopt Cuthill-McKee (C-M) ordering as our ordering scheme.} We empirically discover that C-M ordering \citep{cuthill1969reducing} provides the most compact \tree.\footnote{We provide the results in \cref{subsec:abl}.} Our explanation is that the C-M ordering is specifically designed to align the non-zero elements of a matrix near its diagonal so that there is a higher chance of encountering large submatrices filled only with zeros, which can be efficiently summarized in the \tree representation.

% We also remark that our description assumes the size of the original graph to be the power of $K^{2}$. However, these assumptions can be alleviated by adding ``dummy nodes'' to increase the size of the original graph. This marginally affects the compression ratio since the dummy nodes form submatrices filled with zeros, which can be summarized into a single tree node in the \tree representation. We empirically verify this in \cref{tab:ab_compression}.
\section{Hierarchical Graph Generation with $K^{2}$--trees}\label{sec:method}

\begin{figure}
\centering
  \centering
  \includegraphics[width=0.9\textwidth]{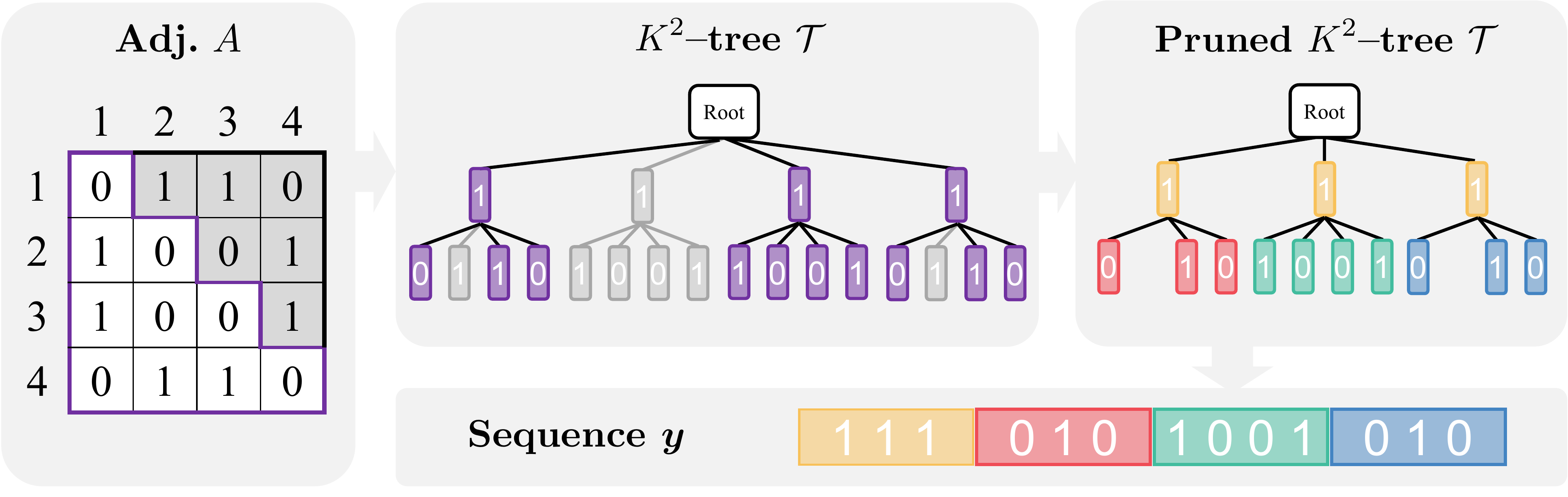}
    \caption{\textbf{Illustration of the sequential representation for \tree.} The shaded parts of the adjacency matrix $A$ and the \tree $\mathcal{T}$ denote redundant parts, which are further pruned, while the purple-colored parts of $A$ and $\mathcal{T}$ denote non-redundant parts. Also, same-colored tree-nodes of pruned \tree are grouped and tokenized into the same colored parts of the sequence $\bm{y}$.} 
\label{fig:seqrep}
\vspace{-.1in}
\end{figure}

In this section, we present our novel method, hierarchical graph generation with \trees (\Algname), exploiting the hierarchical and compact structure of the \tree representation of a graph. In detail, we transform the \tree into a highly compressed sequence through a process involving pruning and tokenization. Subsequently, we employ a Transformer enhanced with tree-based positional encodings, for the autoregressive generation of this compressed sequence.

\subsection{Sequential \tree representation}\label{subsec:seqrep}

Here, we propose an algorithm to flatten the \tree into a sequence, which is essential for the autoregressive generation of the \tree. In particular, we aim to design a sequential representation that is even more compact than the \tree to minimize the number of decisions required for the generation of the \tree. To this end, we propose (1) pruning \tree by removing redundant nodes, (2) flattening the pruned \tree into a sequence, and (3) applying tokenization based on the \tree structure. We provide an illustration of the overall process in \cref{fig:seqrep}.

\textbf{Pruning the \tree.} To obtain the pruned \tree, we identify and eliminate redundant nodes due to the symmetry of the adjacency matrix for undirected graphs. In particular, without loss of generality, such nodes are associated with submatrices positioned above the diagonal since they mirror the counterparts located below the diagonal.

\begin{figure}[t]
\centering
  \includegraphics[width=\linewidth]{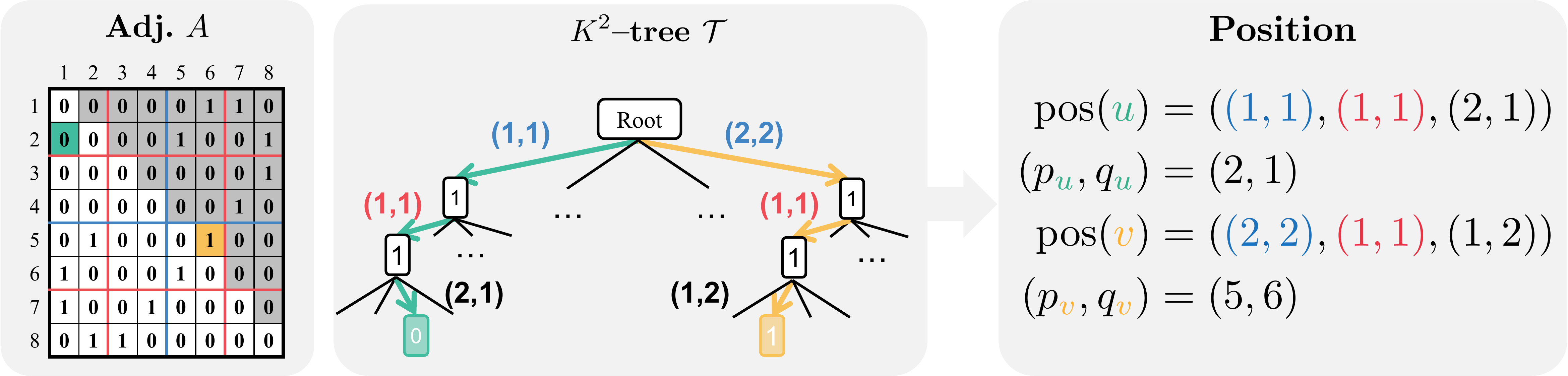}
    \caption{\textbf{Illustration of the tree-node positions of \tree.} The shaded parts of the adjacency matrix denote redundant parts, e.g., $p_{u}<q_{u}$. Additionally, colored elements correspond to tree-nodes of the same color and the same-colored tree-edges signify the root-to-target downward path. Blue and red tuples denote the order in the first and second levels, respectively. The tree node $u$ is non-redundant as $p_u>q_u$ while $v$ is redundant as $p_{v}<q_{v}$.}\label{fig:pos}
  \vspace{-0.2in}
\end{figure}

To this end, we now describe a formula to identify redundant nodes based on the position of a submatrix $A^{(u)}$, tied to a specific node $u$ at depth $L$, within the adjacency matrix $A$. Let $v_{0}, v_{1}, \ldots, v_{L}$ be a sequence of nodes representing a downward path from the root node $r=v_{0}$ to the node $u=v_{L}$. With $(i_{v_{\ell}}, j_{v_{\ell}})$ denoting the order of $v_{\ell}$ among its $K\times K$ siblings, the node position can be represented as $\operatorname{pos}(u)=((i_{v_1}, j_{v_1}), \ldots, (i_{v_L}, j_{v_L}))$. Note that node $u$ at depth $L$ corresponds to an element of $K^{L}\times K^{L}$ partitions of the adjacency matrix $A$. The row and column indexes of the submatrix $A^{(u)}$ are derived as the ${(p_u,q_u)= (\sum_{\ell=1}^{L}K^{L-\ell}(i_{v_{\ell}}-1)+1, \sum_{\ell=1}^{L}K^{L-\ell}(j_{v_{\ell}}-1)+1)}$ as illustrated in \cref{fig:pos}.
As a result, we eliminate any node associated with a submatrix above the diagonal, i.e., we remove node $u$ when $p_u < q_u$.

% \vspace{-0.2in}

Consequently, the pruned \tree maintains only the nodes associated with submatrices devoid of redundant nodes, i.e., those containing elements of the adjacency matrix positioned at the diagonal or below the diagonal. Notably, following this pruning process, the \tree no longer adheres to the structure of a $K \times K$-ary tree. Additionally, consider a non-leaf node $u$ is associated with a submatrix $A^{(u)}$ that includes any diagonal elements of the adjacency matrix $A$. Then the node $u$ possess ${K(K+1)}/{2}$ child nodes after pruning ${K(K-1)}/{2}$ child nodes associated with the redundant submatrices. Otherwise, the non-leaf node $u$ remains associated with $K \times K$ child nodes. Note that our framework can be extended to directed graphs by omitting the pruning process.

\textbf{Flattening and tokenization of the pruned \tree.} Next, we explain how to obtain a sequential representation of the pruned \tree based on flattening and tokenization. Our idea is to flatten a \tree as a sequence of node attributes $\{x_{u}: u \in \mathcal{V}\}$ using breadth-first traversal and then to tokenize the sequence by grouping the nodes that share the same parent node, i.e., sibling nodes.

For this purpose, we denote the sequence of nodes obtained from a breadth-first traversal of non-root nodes in the \tree as $u_{1}, \ldots, u_{|\mathcal{V}|-1}$, and the corresponding sequence of node attributes as $\bm{x}=(x_{1}, \ldots, x_{|\mathcal{V}|-1})$. It is important to note that sibling nodes sharing the same parent appear sequentially in the breadth-first traversal. 

Next, by grouping the sibling nodes, we tokenize the sequence $\bm{x}$. As a result, we obtain a sequence $\bm{y}=(y_{1},\ldots, y_{T})$ where each element is a token representing a group of attributes associated with sibling nodes. For example, the $t$-th token corresponding to a group of $K^{2}$ sibling nodes is represented by $y_{t} = (x_{v_{1,1}},\ldots, x_{v_{K,K}})$ where $v_{1,1},\ldots, v_{K,K}$ share the same parent node $u$. Such tokenization allows representing the whole \tree using $M(\log_{K^2}({N^2}/{M})+O(1))$ space, where $N$ and $M$ denote the number of nodes and edges in the original graph, respectively.

We highlight that the number of elements in each token $y_{t}$ may vary due to the pruned \tree no longer being a $K\times K$-ary tree, as mentioned above. With this in consideration, we generate a vocabulary of $2^{K^{2}}+2^{\nicefrac{K(K+1)}{2}}$ potential configurations for each token $y_{t}$. This vocabulary size is small in practice since we set the value $K$ to be small, e.g., setting $K=2$ induces the size of $24$.

\begin{figure}
\centering
    \includegraphics[width=0.9\linewidth]{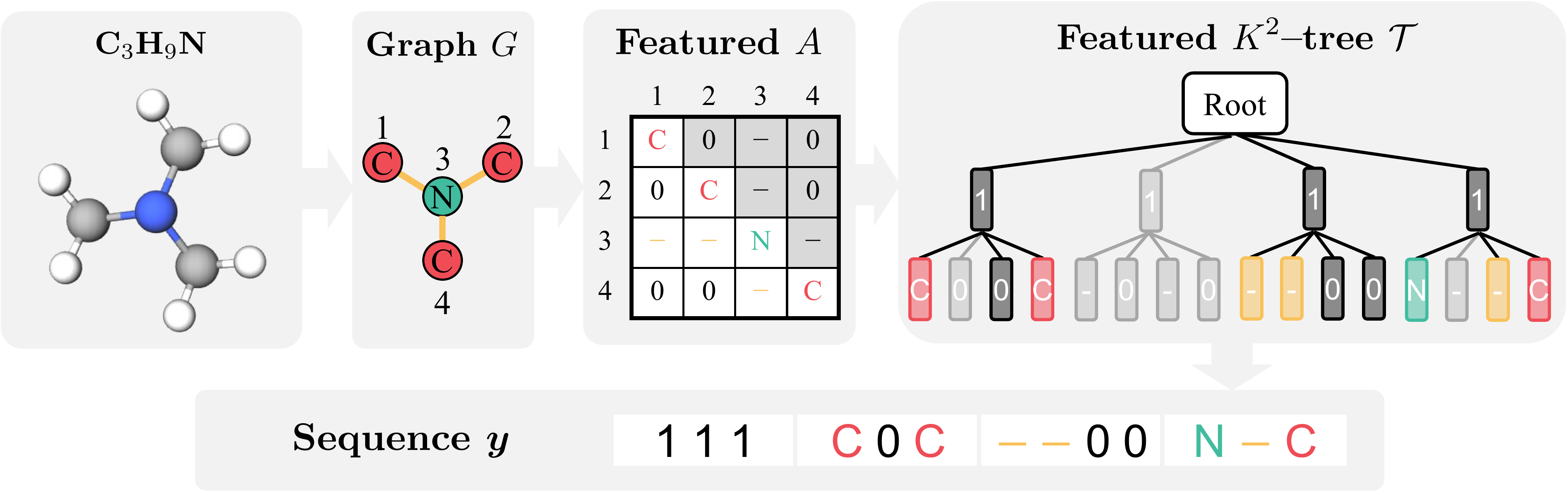}
    \caption{\textbf{An example of featured \tree representation.} The shaded parts of the adjacency matrix and \tree denote the redundant parts. The black-colored tree-nodes denote the normal tree-nodes with binary attributes while other-colored feature elements in the adjacency matrix $A$ denote the same-colored featured tree-nodes and sequence elements. The node features (i.e., C and N) and edge feature (i.e., single bond $-$) of the molecule are represented within the leaf nodes.}
    \label{fig:feature_tree}
    \vspace{-.15in}
\end{figure}
In particular, we remark that a token with ${K(K+1)}/{2}$ elements carries different semantics from another token with $K^{2}$ elements. The former corresponds to a submatrix situated on the adjacency matrix's diagonal, thus indicating connectivity \textit{within} a set of nodes. In contrast, the latter relates to a submatrix illustrating connectivity \textit{between} pairs of node sets. This supports our decision to assign distinct values to a token with ${K(K+1)}/{2}$ elements and another with $K^{2}$ elements, even when the tokens might represent the same combination of node features in the unpruned tree.

\textbf{Generating featured graphs.} We also extend our \Algname to graphs with node and edge-wise features, e.g., molecular graphs. At a high level, we apply our algorithm to the featured adjacency matrix, where each diagonal element corresponds to a node feature and each non-diagonal element corresponds to an edge feature. node attributes of leaf nodes in \tree correspond to node and edge features, while attributes of non-leaf nodes are the same with the non-attributed \trees (i.e., ones and zeros).  See \cref{fig:feature_tree} for an illustration and \cref{appx:attr} for a complete description.

\subsection{Generating \tree with Transformer and \tree positional encoding}\label{subsec:seqgen}

We describe our algorithm to generate the sequence of \tree representation $ \bm{y}=(y_{1},\ldots, y_{T})$. We utilize the masked Transformer \citep{vaswani2017attention} to make predictions on $p_{\theta}(y_{t} | y_{t-1},\ldots, y_{1})$. To improve the model's understanding of the tree structure, we devise a tree-positional encoding. We also offer an algorithm to construct the \tree from the sequence generated by the Transformer.

\textbf{Transformer with \tree positional encoding.} We first introduce the Transformer architecture to parameterize the distribution $p_{\theta}(y_{t} | y_{t-1},\ldots, y_{1})$ for autorgressive generation. Briefly, the model is trained with self-attention, and during inference, it generates the sequence one token at a time, relying on the previously generated sequence. To account for tree structural information, we incorporate tree-positional encodings for each time-step $t$. 

During training, we mask the attention layer to ensure that predictions at each step are not influenced by future tokens of the sequence. The objective function is maximum likelihood, denoted by $\max \log p(\bm{y})$, where $p(\bm{y})=p(y_1)\Pi_{t=2}^T p(y_t|y_{1:t-1})$. This objective aims to maximize the probability of predicting the next token correctly based on the preceding tokens.

For inference, we begin the process with a begin-of-sequence (BOS) token as the input to our trained Transformer decoder. The model then computes the distribution of potential tokens for the next step, denoted by $p(y_t|y_{1:t-1})$, and the next token is sampled from this distribution. This token is appended to the input sequence, and the extended sequence is fed back into the model to generate the subsequent token. This iterative procedure is terminated when a predefined maximum length is reached or an end-of-sequence (EOS) token emerges.

To enhance the input $y_{t}$, we incorporate the positional encoding for $u$.  As outlined in \cref{subsec:seqrep}, the node attributes in $y_{t}$ are associated with child nodes of a particular node $u$. Therefore, the encoding is based on the downward path from the root node $r=v_{0}$ to the node $u=v_{L}$, represented as $(v_{0}, \ldots, v_{L})$. In this context, the order of $v_{\ell}$ amongst its siblings in the non-pruned \tree is denoted as a tuple $(i_{v_{\ell}}, j_{v_{\ell}})$. Subsequently, we further update the input feature $y_{t}$ with positional encoding, which is represented as $\operatorname{PE}(u) = \sum_{\ell=1}^{L}\phi_{\ell}(i_{v_{\ell}}, j_{v_{\ell}})$, where $\phi$ denotes the embedding function that converts the order tuple into vector representations and $((i_{v_{1}}, j_{v_{1}}),\ldots, (i_{v_{L}}, j_{v_{L}}))$ is the sequence of orders of a downward path from $r$ to $u$.

\textbf{Constructing \tree from the sequential representation.} 
We next explain the algorithm to recover a \tree from its sequential representation $\bm{y}$. In particular, we generate the \tree simultaneously with the sequence to incorporate the tree information for each step of the autoregressive generation. The algorithm begins with an empty tree containing only a root node and iteratively expands each ``frontier'' node based on the sequence of the decisions made by the generative model. To facilitate a breadth-first expansion approach, the algorithm utilizes a first-in-first-out (FIFO) queue, which contains node candidates to be expanded. 

To be specific, our algorithm initializes a \tree $\mathcal{T}=(\{r\}, \emptyset)$ with the root node $r$ associated with the node attribute $x_{r}=1$. It also initializes the FIFO queue $\mathcal{Q}$ with $r$. Then at each $t$-th step, our algorithm expands the node $u$ popped from the queue $\mathcal{Q}$ using the token $y_{t}$. To be specific, for each node attribute $x$ in $y_{t}$, our algorithm adds a child node $v$ with $x_{v} = x$. If $x=1$ and the size of $A^{(v)}$ is larger than $1 \times 1$, the child node $v$ is inserted into the queue $Q$. This algorithm is designed to retrieve the pruned tree, which allows the computation of positional data derived from the $y_{t}$ information. 
\section{Experiment}\label{sec:exp}
% In this section, we present the empirical results of graph generation tasks. We evaluate the performance of \Algname against various baseline models on four general and two molecular graph datasets.

\subsection{Generic graph generation}\label{subsec:generic}
\begin{table}
  \caption{\textbf{Generic graph generation performance.} The baseline results are from prior works \citep{jo2022gdss, liao2019efficient, martinkus2022spectre, luo2022fast} or public codes (marked by *). For each metric, the best number is highlighted in \textbf{bold} and the second-best number is \underline{underlined}.}
  \label{tab:main}
  \centering
    \scalebox{0.8}{
    \begin{tabular}{lcccccccccccc}
    \toprule
    % \cmidrule(lr){2-4} \cmidrule(lr){5-7} \cmidrule(lr){8-10} \cmidrule(lr){11-13}
    &\multicolumn{3}{c}{Community-small} & \multicolumn{3}{c}{Planar} & \multicolumn{3}{c}{{Enzymes}} & \multicolumn{3}{c}{Grid} \\
    \cmidrule(lr){2-4} \cmidrule(lr){5-7} \cmidrule(lr){8-10} \cmidrule(lr){11-13}
    & \multicolumn{3}{c}{{$12 \leq |V| \leq 20$}} & \multicolumn{3}{c}{{$|V|=64$}} & \multicolumn{3}{c}{{$10 \leq |V| \leq 125$}} & \multicolumn{3}{c}{{$100 \leq |V| \leq 400$}} \\
    \cmidrule(lr){2-4} \cmidrule(lr){5-7} \cmidrule(lr){8-10} \cmidrule(lr){11-13}
    Method & Deg. & Clus. & Orb. & Deg. & Clus. & Orb. & Deg. & Clus. & Orb. & Deg. & Clus. & Orb. \\
    \midrule
    GraphVAE  & 0.350 & 0.980 & 0.540 & - & - & - & 1.369 & 0.629 & 0.191 & 1.619 & \textbf{0.000} & 0.919 \\
    GraphRNN & 0.080 & 0.120 & 0.040 & 0.005 & 0.278 & 1.254 & 0.017 & 0.062 & 0.046 & 0.064 & 0.043 & 0.021 \\
    GNF  & 0.200 & 0.200 & 0.110 & - & - & - & - & - & - & - & - & - \\
    GRAN*  & 0.005 & 0.142 & 0.090 & \underline{0.001} & {0.043} & \underline{0.001} & 0.023 & 0.031 & 0.169 & \underline{0.001} & \underline{0.004} & \underline{0.002} \\
    EDP-GNN  & 0.053 & 0.144 & 0.026 & - & - & - & 0.023 & 0.268 & 0.082 & 0.455 & 0.238 & 0.328 \\
    GraphGen* & 0.075 & 0.065 & 0.014 & 1.762 & 1.423 & 1.640 & 0.146 & 0.079 & 0.054 & 1.550 & 0.017 & 0.860 \\
    GraphAF  & 0.180 & 0.200 & 0.020 & - & - & - & 1.669 & 1.283 & 0.266 & - & - & - \\
    GraphDF  & 0.060 & 0.120 & 0.030 & - & - & - & 1.503 & 1.283 & 0.266 & - & - & - \\
    SPECTRE  & - & - & - & 0.010 & 0.067 & 0.010 & - & - & - & - & - & - \\
    GDSS  & 0.045 & 0.086 & 0.007 & 0.250 & 0.393 & 0.587 & 0.026 & 0.061 & \underline{0.009} & 0.111 & 0.005 & 0.070 \\
    DiGress* & 0.012 & 0.025 & \underline{0.002} & \textbf{0.000} & \underline{0.002} & 0.008 & \underline{0.011} & \underline{0.039} & 0.010 & 0.016 & \textbf{0.000} & 0.004 \\
    GDSM & \underline{0.011} & \underline{0.015} & \textbf{0.001} & - & - & - & 0.013 & 0.088 & 0.010 & 0.002 & \textbf{0.000} & \textbf{0.000} \\
    \midrule
    \Algname (ours) & \textbf{0.001} & {\textbf{0.006}} & 0.003  & \textbf{0.000} & \textbf{0.001} & \textbf{0.000} & \textbf{0.005} & \textbf{0.017} & \textbf{0.000} & \textbf{0.000} & \textbf{0.000} & \textbf{0.000} \\
    \bottomrule
    \end{tabular}}
    \vspace{-.1in}
\end{table}

\begin{figure}[t!]
    \centering
    \begin{subfigure}[t]{0.19\linewidth}
    \centering
      \includegraphics[width=\linewidth]{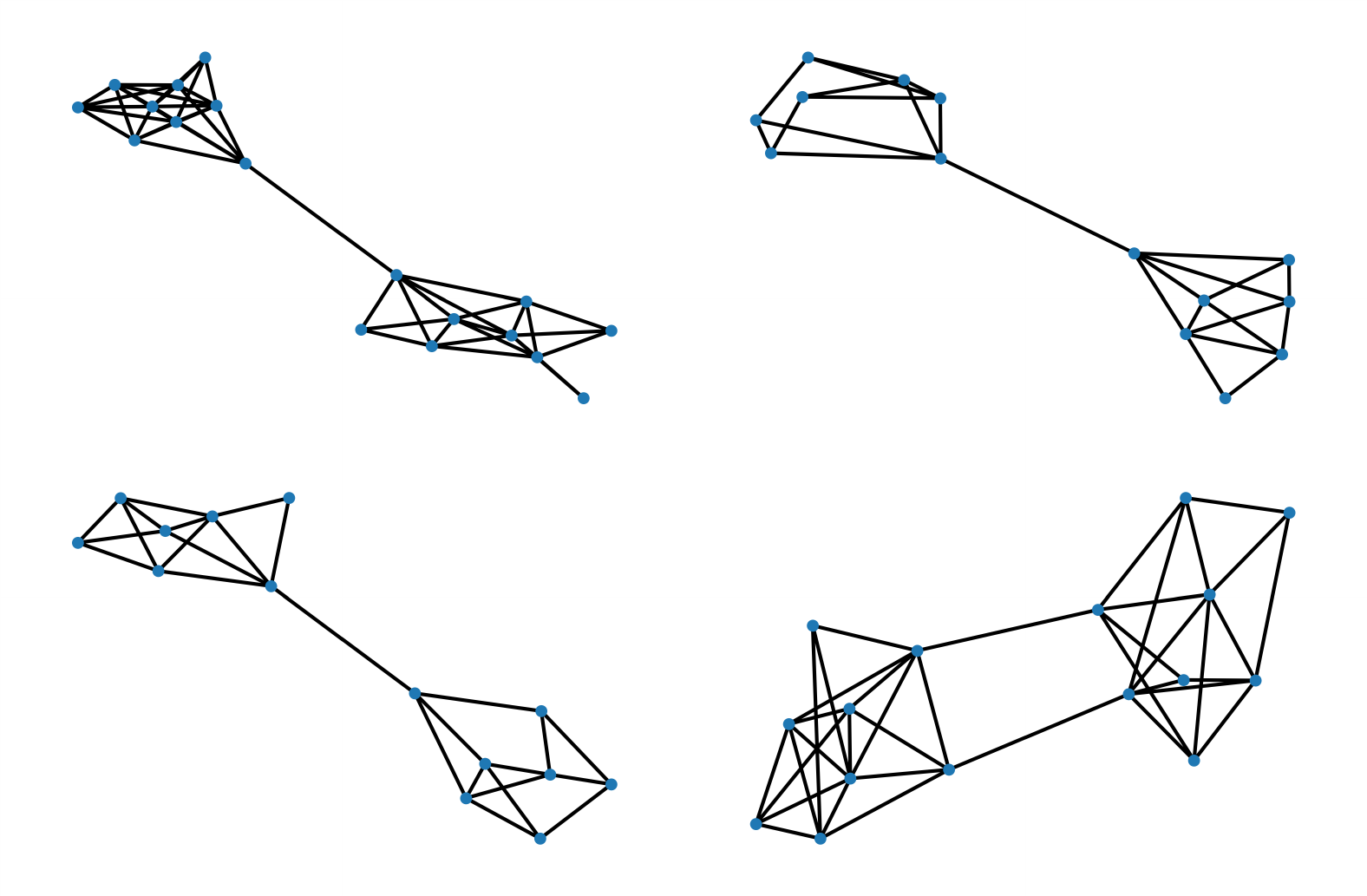}
      %\subcaption{Train}
      %\label{subfig:com_train}
    \end{subfigure}
    \hfill
    \begin{subfigure}[t]{0.19\linewidth}
    \centering
      \includegraphics[width=\linewidth]{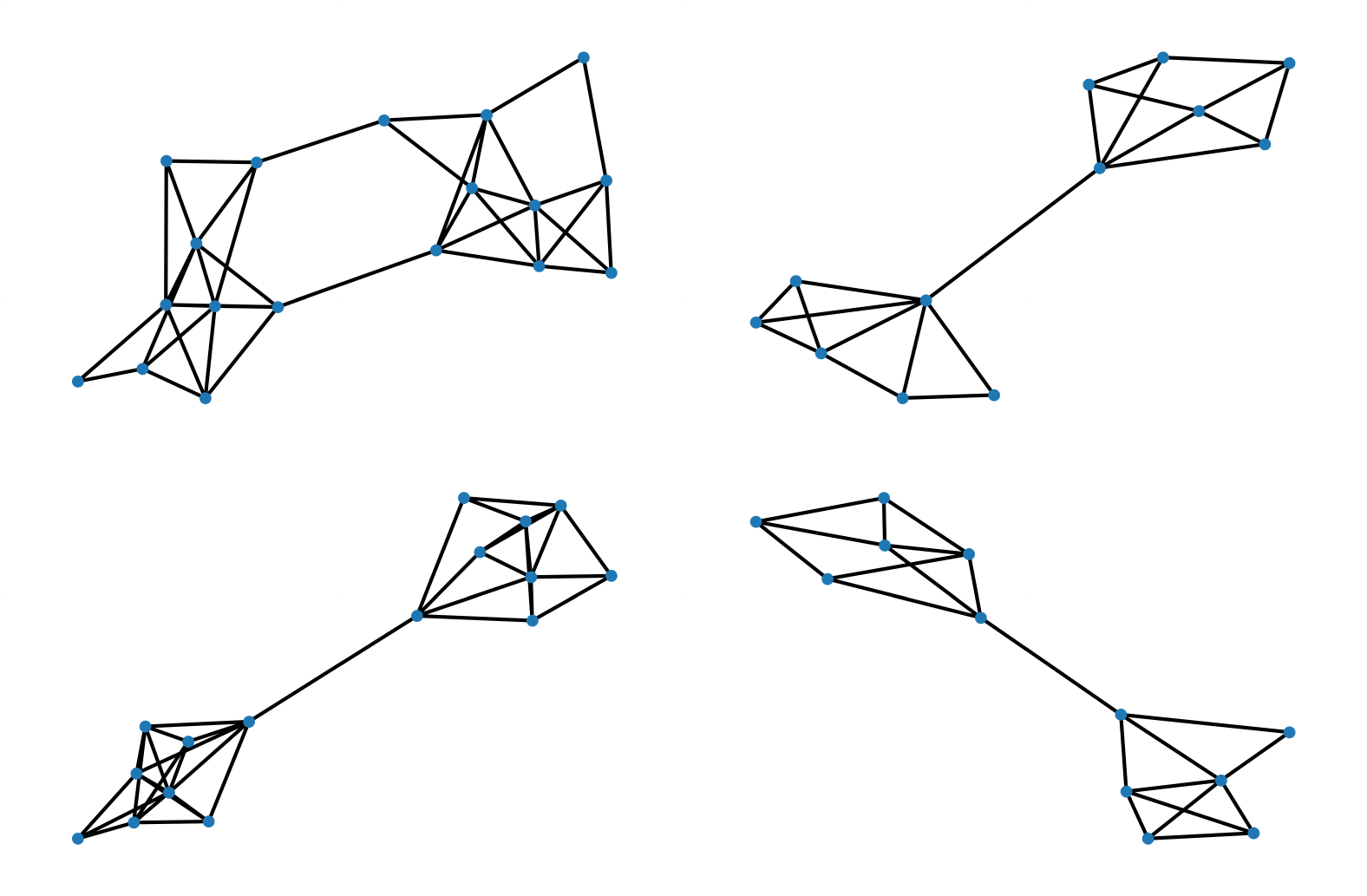}
      %\subcaption{GraphGen}
      %\label{subfig:com_train}
    \end{subfigure}
    \hfill
    \begin{subfigure}[t]{0.19\linewidth}
    \centering
      \includegraphics[width=\linewidth]{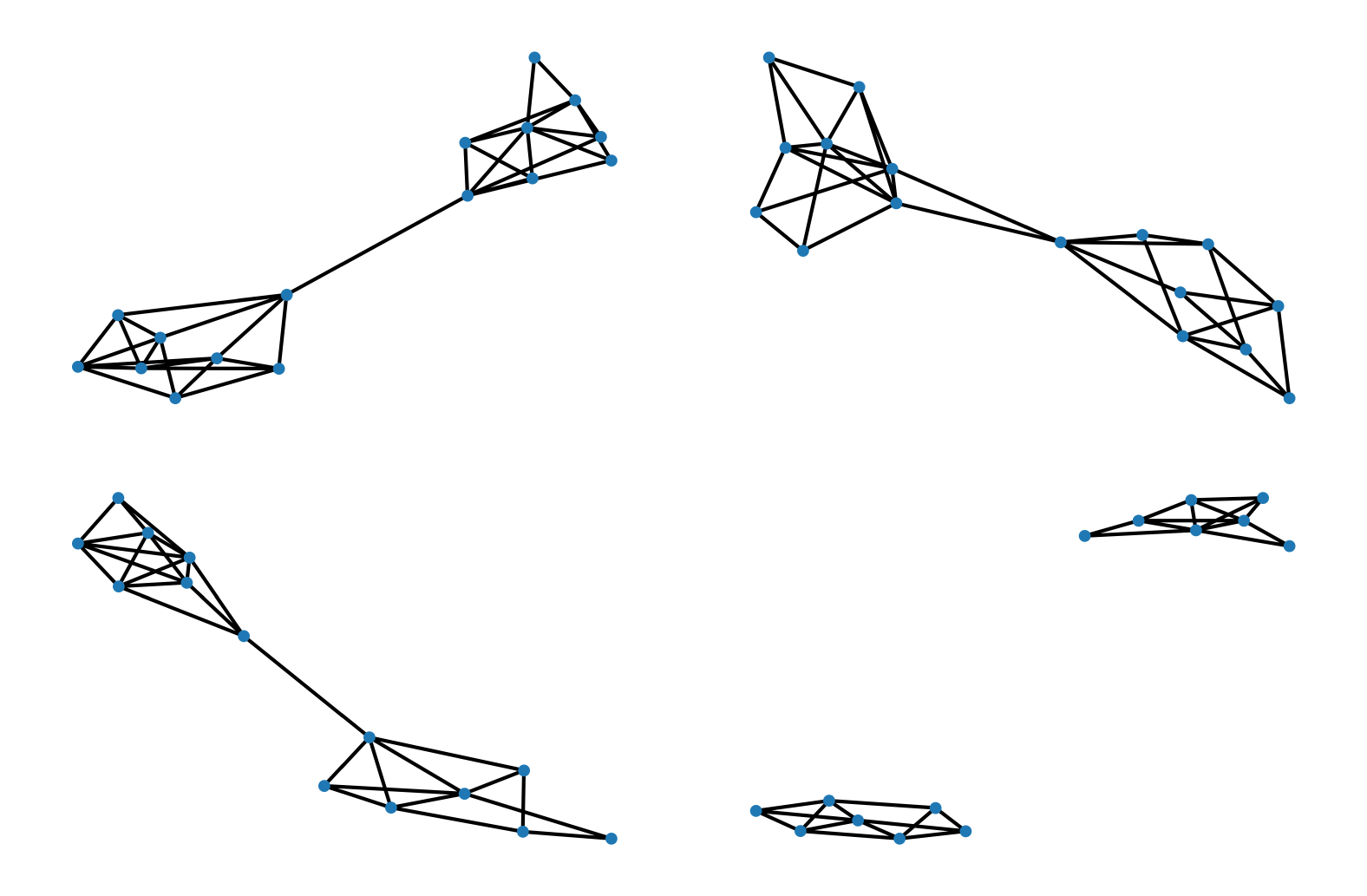}
      %\subcaption{GDSS}
      %\label{subfig:com_train}
    \end{subfigure}
    \hfill
    \begin{subfigure}[t]{0.19\linewidth}
    \centering
      \includegraphics[width=\linewidth]{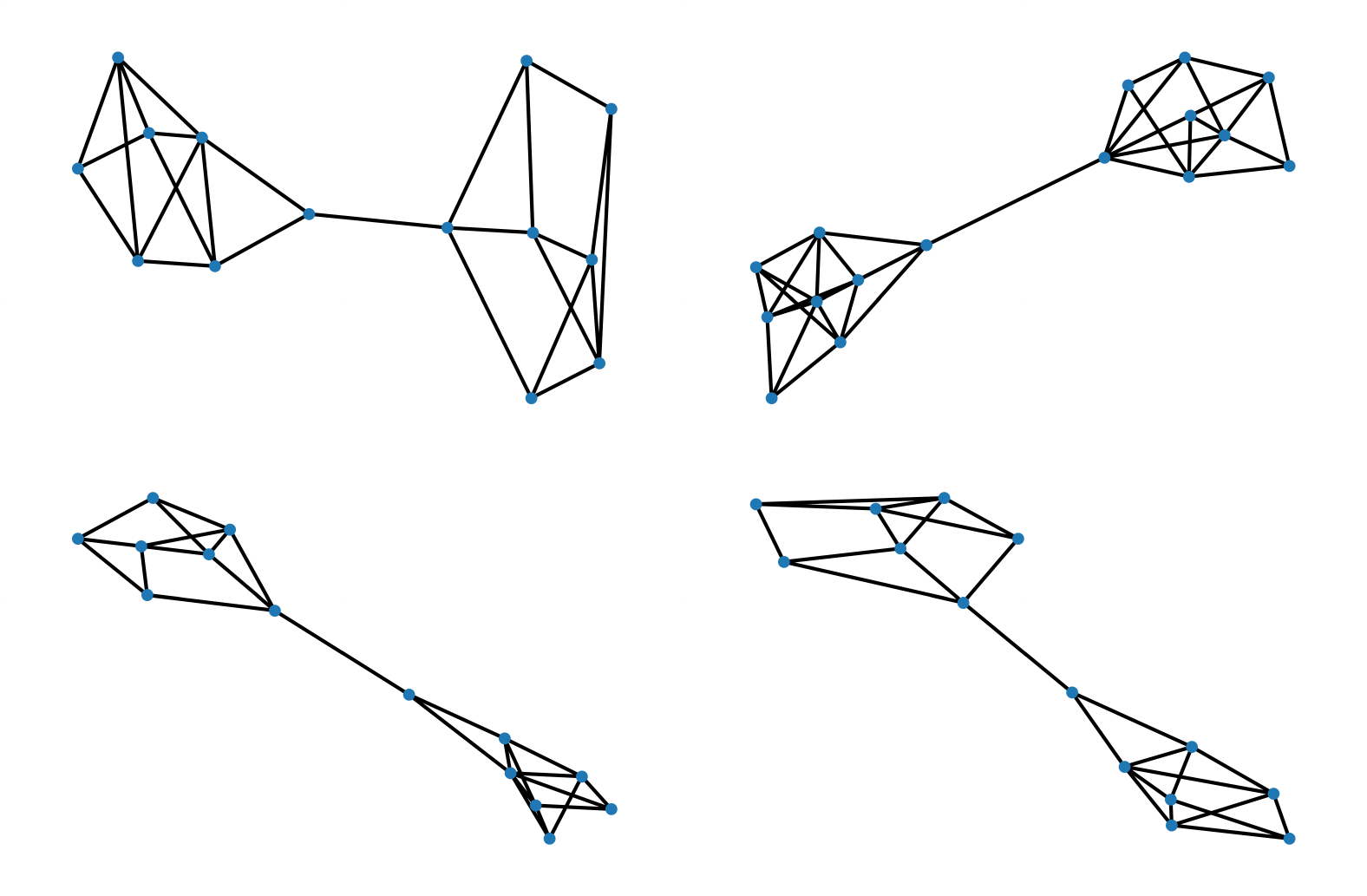}
      %\subcaption{DiGress}
      %\label{subfig:com_train}
    \end{subfigure}
    \hfill
    \begin{subfigure}[t]{0.19\linewidth}
    \centering
      \includegraphics[width=\linewidth]{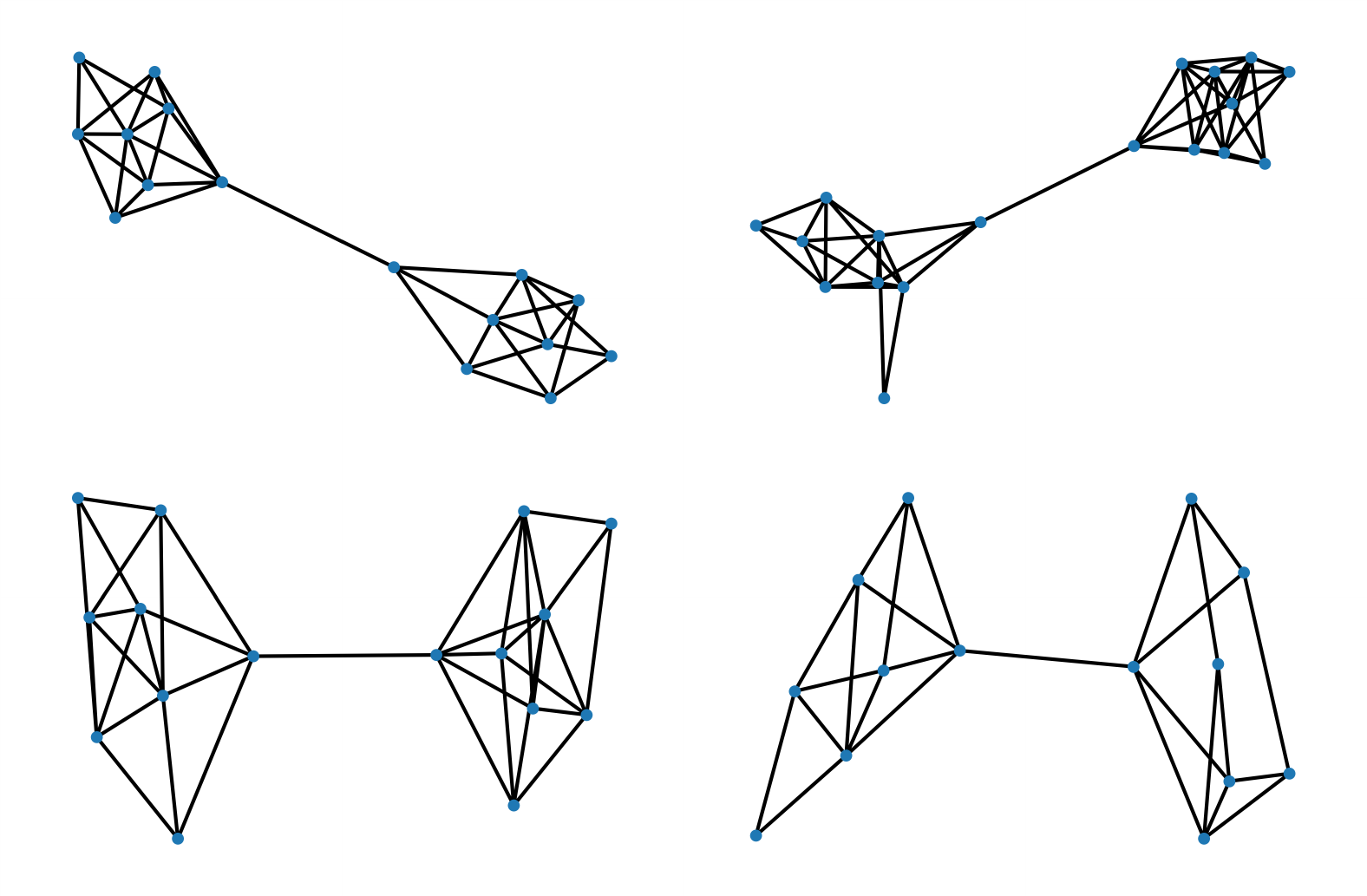}
      %\subcaption{\Algname (ours)}
      %\label{subfig:com_train}
    \end{subfigure}
    \vskip\baselineskip
    % \begin{subfigure}[t]{0.19\linewidth}
    % \centering
    %   \includegraphics[width=\linewidth]{figure/generated_samples/GDSS_enz-train.pdf}
    %   %\subcaption{Train}
    %   %\label{subfig:com_train}
    % \end{subfigure}
    % \hfill
    % \begin{subfigure}[t]{0.19\linewidth}
    % \centering
    %   \includegraphics[width=\linewidth]{figure/generated_samples/GDSS_enz-graphgen.pdf}
    %   %\subcaption{GraphGen}
    %   %\label{subfig:com_train}
    % \end{subfigure}
    % \hfill
    % \begin{subfigure}[t]{0.19\linewidth}
    % \centering
    %   \includegraphics[width=\linewidth]{figure/generated_samples/GDSS_enz-gdss.pdf}
    %   %\subcaption{GDSS}
    %   %\label{subfig:com_train}
    % \end{subfigure}
    % \hfill
    % \begin{subfigure}[t]{0.19\linewidth}
    % \centering
    %   \includegraphics[width=\linewidth]{figure/generated_samples/GDSS_enz-digress.pdf}
    %   %\subcaption{DiGress}
    %   %\label{subfig:com_train}
    % \end{subfigure}
    % \hfill
    % \begin{subfigure}[t]{0.19\linewidth}
    % \centering
    %   \includegraphics[width=\linewidth]{figure/generated_samples/GDSS_enz-gcg.pdf}
    %   %\subcaption{\Algname (ours)}
    %   %\label{subfig:com_train}
    % \end{subfigure}
    % \vskip\baselineskip
    \begin{subfigure}[t]{0.19\linewidth}
    \centering
      \includegraphics[width=\linewidth]{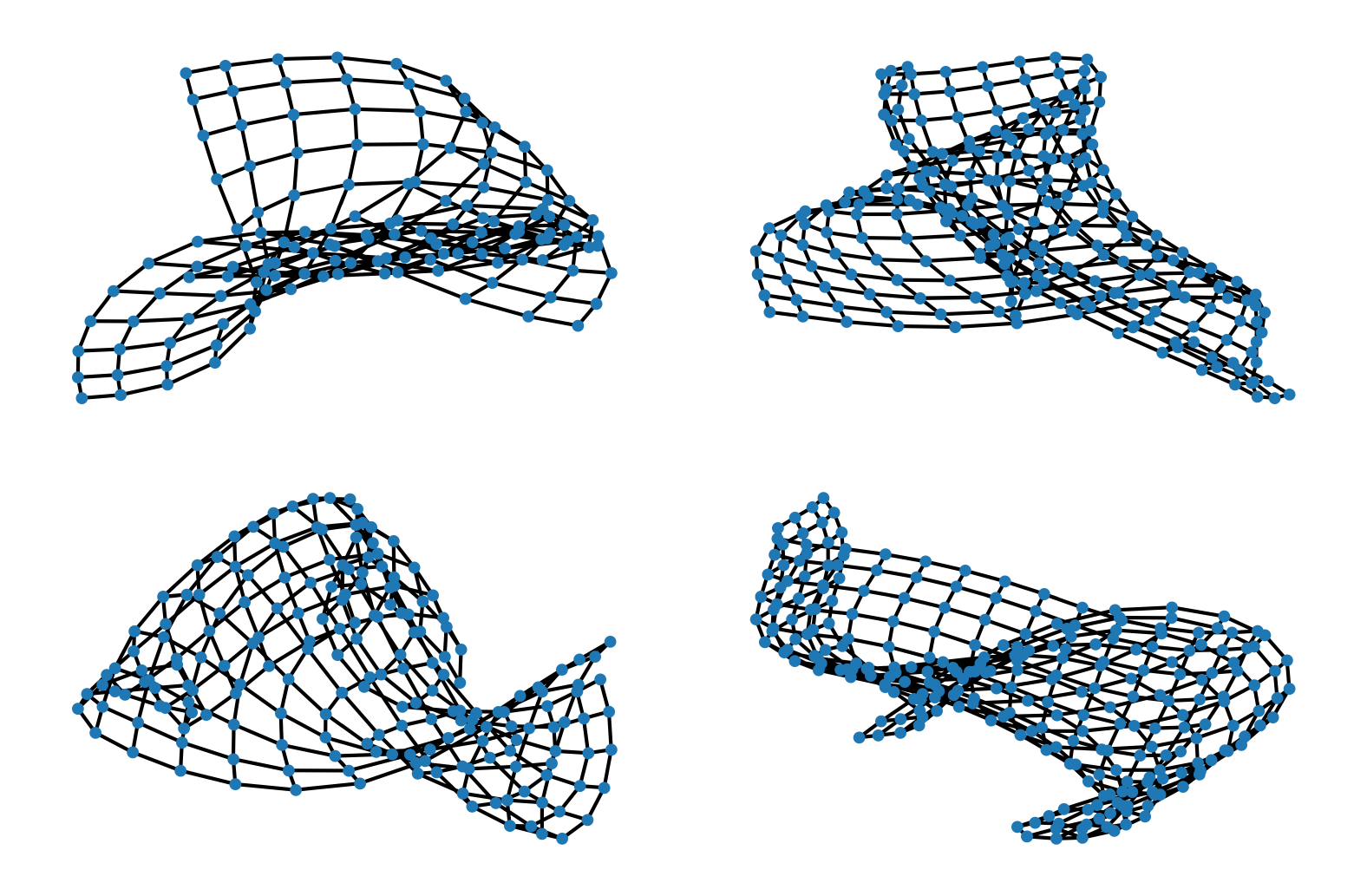}
      \subcaption{Train}
      \label{subfig:com_train}
    \end{subfigure}
    \hfill
    \begin{subfigure}[t]{0.19\linewidth}
    \centering
      \includegraphics[width=\linewidth]{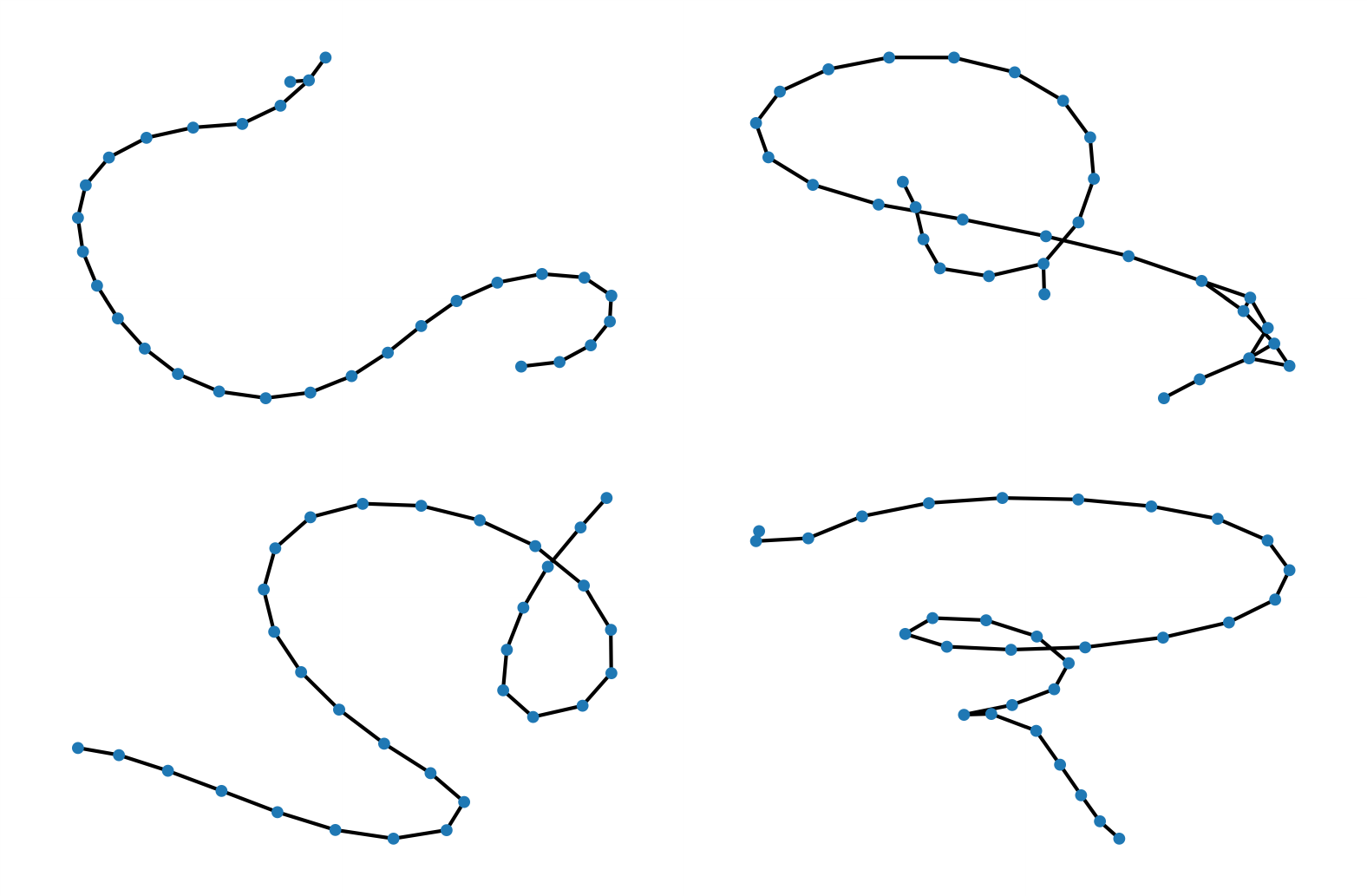}
      \subcaption{GraphGen}
      \label{subfig:com_train}
    \end{subfigure}
    \hfill
    \begin{subfigure}[t]{0.19\linewidth}
    \centering
      \includegraphics[width=\linewidth]{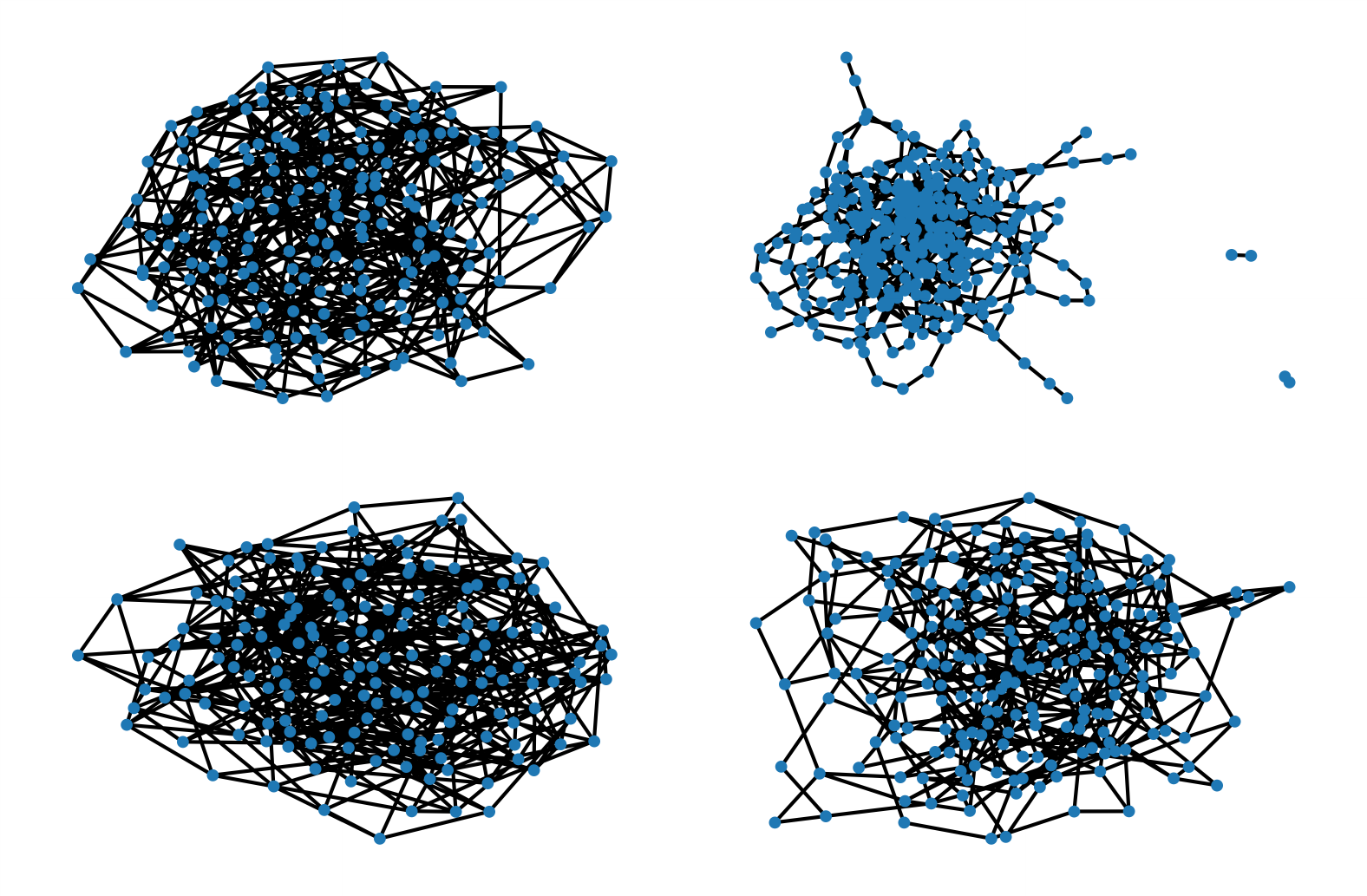}
      \subcaption{GDSS}
      \label{subfig:com_train}
    \end{subfigure}
    \hfill
    \begin{subfigure}[t]{0.19\linewidth}
    \centering
      \includegraphics[width=\linewidth]{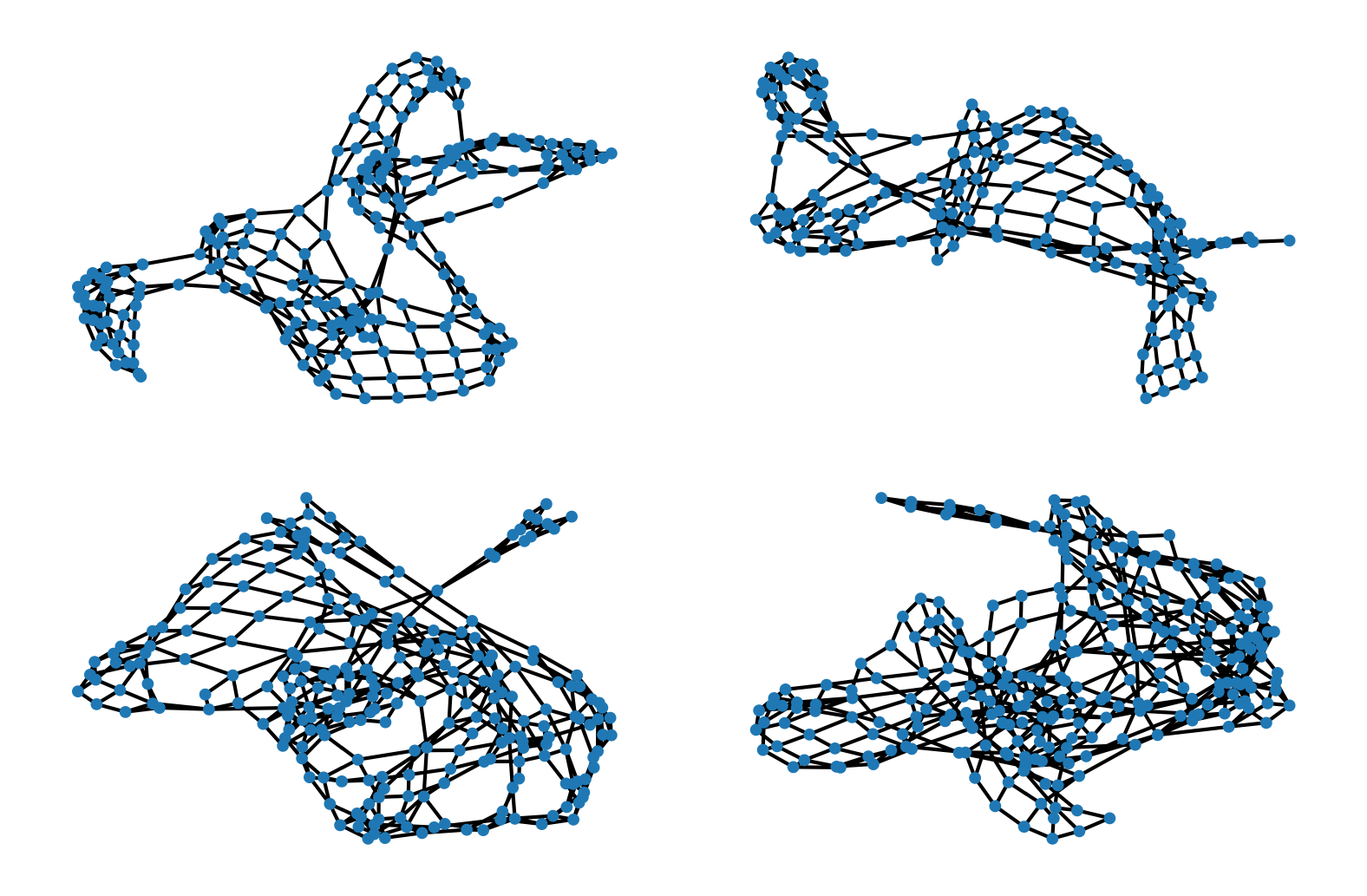}
      \subcaption{DiGress}
      \label{subfig:com_train}
    \end{subfigure}
    \hfill
    \begin{subfigure}[t]{0.19\linewidth}
    \centering
      \includegraphics[width=\linewidth]{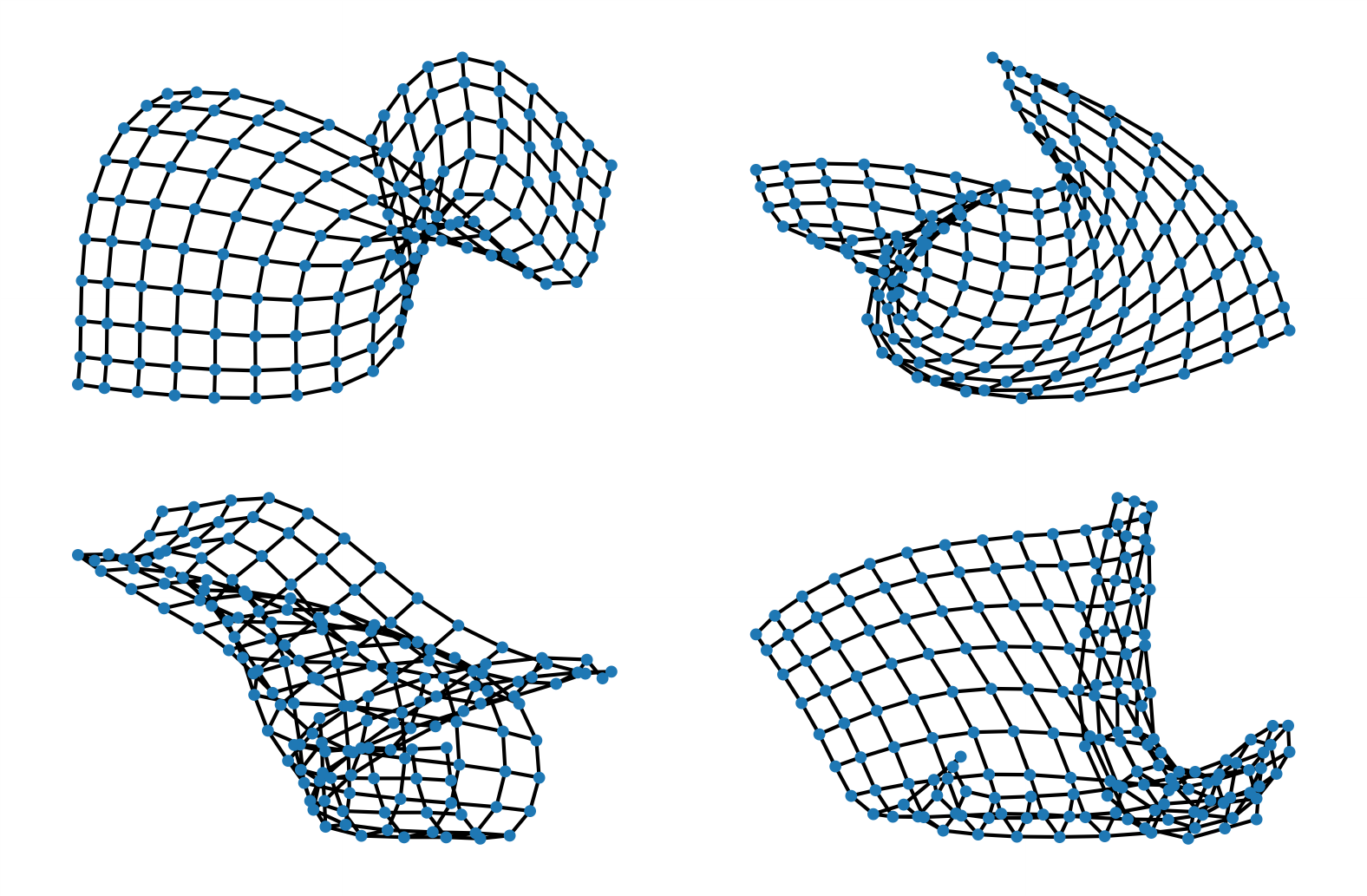}
      \subcaption{\Algname (ours)}
      \label{subfig:com_train}
    \end{subfigure}

    \caption{\textbf{Generated samples for Community-small (top), and Grid (bottom) datasets.}}\label{fig:exp_main}
  \vspace{-.1in}
\end{figure}

\textbf{Experimental setup.} We first validate the general graph generation performance of our \Algname on four popular graph benchmarks: (1) \textbf{Community-small}, 100 community graphs, (2) \textbf{Planar}, 200 planar graphs, (3) \textbf{Enzymes} \citep{schomburg2004brenda}, 587 protein tertiary structure graphs, and (4) \textbf{Grid}, 100 2D grid graphs. 
%We use the same split with the experiments conducted by Martinkus et al.~\citep{martinkus2022spectre} and Jo et al.~\citep{jo2022gdss}. 
Following baselines, we adopt maximum mean discrepancy (MMD) to compare three graph property distributions between generated graphs and test graphs: degree (\textbf{Deg.}), clustering coefficient (\textbf{Clus.}), and 4-node-orbit counts (\textbf{Orb.}). \textcolor{black}{We conduct all the experiments using a single RTX 3090 GPU.} The detailed descriptions of our experimental setup are in \cref{appx:exp}.

\textbf{Baselines.} We compare our \Algname with twelve graph generative models: GraphVAE \citep{simonovsky2018graphvae},  GraphRNN \citep{you2018graphrnn},  GNF \cite{liu2019gnf}, GRAN \citep{liao2019efficient}, EDP-GNN \citep{niu2020edpgnn}, GraphGen \citep{goyal2020graphgen}, GraphAF \citep{shi2020graphaf}, GraphDF \citep{luo2021graphdf}, SPECTRE \citep{martinkus2022spectre}, GDSS \citep{jo2022gdss}, DiGress \citep{vignac2022digress}, and GDSM \citep{luo2022fast}. A detailed implementation description is in \cref{appx:impl}.

\textbf{Results.} \cref{tab:main} shows the experimental results. We observe that \Algname outperforms all baselines on all datasets. Note that our model consistently outperforms all baselines regardless of the graph sizes, indicating better generalization performance across various environments. In particular, we observe how the performance of \Algname is extraordinary for Grid. We hypothesize that \Algname better captures the hierarchical structure and repetitive local connectivity of the grid graphs than the other baselines. We also provide visualizations of the generated graphs in \cref{fig:exp_main}.

\subsection{Molecular graph generation}\label{subsec:mol}

\begin{table}
  \caption{\textbf{Molecular graph generation performance.} The baseline results are from prior works \citep{jo2022gdss, luo2022fast} or obtained by running the open-source codes (denoted by *). The best results are highlighted in \textbf{bold} and the second best results are \underline{underlined}.}
  \label{tab:mol}
  \centering
  \scalebox{0.8}{
    \begin{tabular}{lcccccccccc}
    \toprule & \multicolumn{5}{c}{ QM9 } & \multicolumn{5}{c}{ ZINC250k } \\ 
    \cmidrule(lr){2-6} \cmidrule(lr){7-11}
    Method & Val. $\uparrow$ & NSPDK $\downarrow$ & FCD $\downarrow$ & Uniq. $\uparrow$ & Nov. $\uparrow$ & Val. $\uparrow$ & NSPDK $\downarrow$ & FCD $\downarrow$ & Uniq. $\uparrow$ & Nov. $\uparrow$ \\
    \midrule    
    EDP-GNN & 47.52 & 0.005 & 2.68 & \textbf{99.25} & 86.58 & 82.97 & 0.049 & 16.74 & 99.79 & \textbf{100}  \\
    MoFlow & 91.36 & 0.017 & 4.47 & \underline{98.65} & 94.72 & 63.11 & 0.046 & 20.93 & \textbf{99.99} & \textbf{100} \\
    GraphAF & 74.43 & 0.020 & 5.27 & 88.64 & 86.59 &  68.47 & 0.044 & 16.02 & 98.64 & \textbf{100} \\
    GraphDF & 93.88 & 0.064 & 10.93 & 98.58 & \textbf{98.54} & 90.61 & 0.177 & 33.55 & 99.63 & \textbf{100} \\
    GraphEBM & 8.22 & 0.030 & 6.14 & 97.90 & 97.01 &  5.29 & 0.212 & 35.47 & 98.79 & \textbf{100} \\
    GDSS & 95.72 & 0.003 & 2.9 & 98.46 & 86.27 & \underline{97.01} & \underline{0.019} & \underline{14.66} & 99.64 & \textbf{100} \\
    DiGress* & 99.01 & \underline{0.001} & \textbf{0.25} & 96.34 & 35.46 & \textbf{\textbf{100}} & 0.042 & 16.54 & \underline{99.97} & \textbf{100}  \\
    GDSM & \textbf{99.90} & 0.003 & 2.65 & - & - & 92.70 & 0.017 & 12.96 & - & -  \\
    \midrule
    \Algname (ours) & \underline{99.22} & \textbf{0.000} & \underline{0.40} & 95.65 & 24.01 & {92.87} & \textbf{0.001} & \textbf{1.93} & \underline{99.97} & \underline{99.83} \\
    \bottomrule
    \end{tabular}
    }
    \vspace{-0.2in}
\end{table}
\textbf{Experimental setup.} To test the ability of \Algname on featured graphs, we further conduct an evaluation of molecule generation tasks. We use two molecular datasets: QM9 \citep{ramakrishnan2014quantum} and ZINC250k \citep{irwin2012zinc}. Following the previous work \citep{jo2022gdss}, we evaluate 10,000 generated molecules using \textcolor{black}{five} metrics: (a) validity (\textbf{Val.}), (b) neighborhood subgraph pairwise distance kernel (\textbf{NSPDK}), (c) Frechet ChemNet Distance (\textbf{FCD}), (d) uniqueness (\textbf{Uniq.}), and (e) novelty (\textbf{\textcolor{black}{Nov.}}). \textcolor{black}{Note that NSPDK and FCD are measured between the generated samples and the test set. The validity, uniqueness, and novelty metrics are measured within the generated samples.}

\textbf{Baselines.} We compare \Algname with eight deep graph generative models: EDP-GNN \citep{niu2020edpgnn}, MoFlow \citep{zang2020moflow}, GraphAF \citep{shi2020graphaf}, GraphDF \citep{luo2021graphdf}, GraphEBM \citep{liu2021graphebm}, GDSS \citep{jo2022gdss}, DiGress \citep{vignac2022digress}, and GDSM\citep{luo2022fast}. We provide a detailed implementation description in \cref{appx:impl}.

\textbf{Results.} The experimental results are reported in \cref{tab:mol}. We observe that \Algname showed competitive results on all the baselines on most of the metrics. The results suggest that the model can generate chemically valid features, i.e., atom types, accordingly, along with the structure of the graphs. In particular, for the ZINC250k dataset, we observe a large gap between our method and the baselines in NSPDK and FCD scores while showing competitive performance in the other metrics. Since FCD and NSPDK measure the similarity between molecular features and subgraph structures, respectively, \Algname can generate similar features and subgraphs observed in the real molecules.

\subsection{Ablation studies}\label{subsec:abl}

\begin{figure*}[t]
    \begin{minipage}{0.5\linewidth}
        \begin{tabular}{C{1.9cm}C{1cm}C{1cm}C{1cm}C{1cm}}
            \toprule 
            Time (sec) & Comm. & Planar & Enzymes & Grid \\
            \midrule
            GRAN  & 3.51 & 5.40 & 3.99 & 14.68 \\
            GDSS & 0.54 & 8.85 & 1.09 & 25.90 \\
            DiGress & 0.34 & 3.29 & 1.29 & 45.41 \\
            \Algname (ours) & \textbf{0.03} & \textbf{0.58} & \textbf{0.09} & \phantom{0}\textbf{8.16} \\
            \bottomrule
        \end{tabular}
        \label{tab:ab_time}
        
        \begin{tabular}{C{1.9cm}C{1cm}C{1cm}C{1cm}C{1cm}}
            \toprule 
            Method & Comm. & Planar & Enzymes & Grid \\
            \midrule
            BFS  & 0.534 & 0.201 & 0.432 & 0.048 \\
            DFS & 0.619 & 0.204 & 0.523 & 0.064 \\
            C-M & \textbf{0.508} & \textbf{0.195} & \textbf{0.404} & \textbf{0.045} \\
            \bottomrule
        \end{tabular}
        \label{tab:ab_compression}
    \end{minipage}
    \hfill
    \begin{minipage}{0.43\linewidth}
        \centering
        \includegraphics[width=0.9\linewidth]{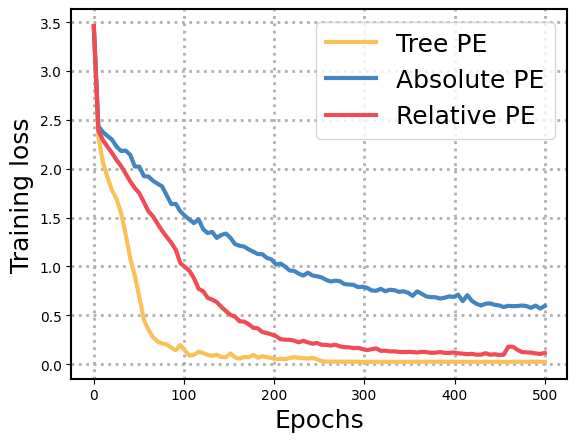}
    \end{minipage}
    \caption{\textbf{(Upper left) Inference time to generate a single graph. (Lower left) Average compression ratio on various node orderings. (Right) Training loss on different positional encodings.}}\label{fig: ablation}
\end{figure*}
\vspace{-0.1in}

\begin{table}
  \caption{\textbf{Ablation study for algorithmic components of \Algname.}}
  \vspace{-.1in}
  \label{tab:ab_rep}
  \centering
  \scalebox{0.8}{
    \begin{tabular}{ccccccccccccc}
    \toprule
    & \multicolumn{2}{c}{} & \multicolumn{3}{c}{{Community-small}} & \multicolumn{3}{c}{{Planar}} & \multicolumn{3}{c}{Enzymes} \\
    \cmidrule(lr){4-6} \cmidrule(lr){7-9} \cmidrule(lr){10-12} 
    Group & TPE & Prune & Degree & Cluster. & Orbit & Degree & Cluster. & Orbit & Degree & Cluster. & Orbit \\
    \midrule
    %Pure transformer & - & - & - & - & - & - & - & - & - & - & - & - \\
    \xmark & \xmark & \xmark & 0.072 & 0.199 & 0.080 & 0.346 & 1.824 & 1.403 & 0.050 & 0.060 & 0.021 \\
    \cmark & \xmark & \xmark  & 0.009 & 0.105 & \textbf{0.001} & \underline{0.003} & \textbf{0.001} & \underline{0.002} & \underline{0.005} & 0.022 & 0.007 \\
    \cmark & \cmark & \xmark  & \underline{0.002} & \underline{0.028} & {\textbf{0.001}} & \underline{0.003} & \textbf{0.001} & \underline{0.002} & \textbf{0.002} & \underline{0.020} & \underline{0.002} \\
    \cmark & \cmark & \cmark & \textbf{0.001} & {\textbf{0.006}} & \underline{0.003} & \textbf{0.000} & \textbf{0.001} & \textbf{0.000} & \underline{0.005} & \textbf{0.017} & \textbf{0.000} \\
    \bottomrule
  \end{tabular}
  }
\end{table}
% \vspace{-0.1in}
\textbf{Time complexity.} We conduct experiments to measure the inference time of the proposed algorithm. The results are presented in the \textcolor{black}{upper left} table of \cref{fig: ablation}, where we report the time to generate a single sample. We can observe that \Algname generates a graph faster than the others due to the simplified representation.

\textbf{Adjacency matrix orderings.} It is clear that the choice of node ordering influences the size of \tree. We validate our choice of Cuthill-McKee (C-M) ordering \citep{cuthill1969reducing} by comparing its compression ratio to other node orderings: breadth-first search (BFS) and depth-first search (DFS). The compression ratio is defined as the number of elements in \tree divided by $N^2$. In the left below table of \cref{fig: ablation}, we present the compression ratios for each node ordering. One can observe that C-M ordering shows the best ratio in all the datasets compared to others. 

\textbf{Positional encoding.} In this experiment, we assess the impact of various positional encodings in our method. We compare our tree positional encoding (TPE) to absolute positional encoding (APE) \citep{vaswani2017attention} and relative positional encoding (RPE) \citep{shaw2018self} on the Planar dataset. Our findings, as presented in the right figure of \cref{fig: ablation}, demonstrate that TPE outperforms other positional encodings with faster convergence of training loss. These observations highlight the importance of appropriate positional encoding for generating high-quality graphs.

\textbf{Ablation of algorithmic components.} We introduce three components to enhance the performance of \Algname: grouping into tokens (Group), incorporating tree positional encoding (TPE), and pruning the \tree (Prune). To verify the effectiveness of each component, we present the experimental results for our method with incremental inclusion of these components. The experimental results are reported in \cref{tab:ab_rep}. The results demonstrate the importance of each component in improving graph generation performance, with grouping being particularly crucial, thereby validating the significance of our additional components to the sequential \tree representation. 

\FloatBarrier 

% We also remark that our description assumes the size of the original graph to be the power of $K^{2}$. However, these assumptions can be alleviated by adding ``dummy nodes'' to increase the size of the original graph. This marginally affects the compression ratio since the dummy nodes form submatrices filled with zeros, which can be summarized into a single tree node in the \tree 

% \textbf{Removing redundancy}
% \input{table/tab_ab_red}

\FloatBarrier 

\section{Conclusion}\label{sec:conclusion}

In this paper, we presented a novel \tree-based graph generative model (\Algname) which enables a compact, hierarchical, and domain-agnostic generation. Our experimental evaluation demonstrated state-of-the-art performance across various graph datasets. An interesting avenue for future work is the broader examination of other graph representations to graph generation, e.g., a plethora of representations \citep{boldi2009permuting,larsson2000off}. 
%\textcolor{red}{In addition, conditional graph generation with \tree representation could be an interesting venue.}

%Furthermore, we showcased the advantages of our method in capturing hierarchical graph structures and generating large graphs with competitive performance when compared to one-shot models. We leave various applications of \tree graph representation for other generative models such as diffusion models as future work.

%Our key idea was to leverage the benefits of the \tree data structure for graph generation tasks. We incorporated transformer architecture with tree positional encoding for \tree graph generation, which allowed our model to capture local and global dependencies between tree nodes effectively.

\clearpage

\textbf{Reproducibility} All experimental code related to this paper is available at \url{https://github.com/yunhuijang/HGGT}. Detailed insights regarding the experiments, encompassing dataset and model specifics, are available in \cref{sec:exp}. For intricate details like hyperparameter search, consult \cref{appx:exp}. In addition, the reproduced dataset for each baseline is in \cref{appx:impl}.

\textbf{Acknowledgements} This work partly was supported by Institute of Information \& communications Technology Planning \& Evaluation (IITP) grant funded by the Korea government (MSIT) (No. IITP-2019-0-01906, Artificial Intelligence Graduate School Program (POSTECH)), the National Research Foundation of Korea (NRF) grant funded by the Korea government (MSIT) (No. 2022R1C1C1013366), Basic Science Research Program through the National Research Foundation of Korea (NRF) funded by the Ministry of Education (2022R1A6A1A0305295413, 2021R1C1C1011375), and the Technology Innovation Program (No. 20014926, Development of BIT Convergent AI Architecture, Its Validation and Candidate Selection for COVID19 Antibody, Repositioning and Novel Synthetic Chemical Therapeutics) funded by the Ministry of Trans, Industry \& Energy (MOTIE, Korea).

\bibliography{iclr2024_conference}

\begin{thebibliography}{46}
\providecommand{\natexlab}[1]{#1}
\providecommand{\url}[1]{\texttt{#1}}
\expandafter\ifx\csname urlstyle\endcsname\relax
  \providecommand{\doi}[1]{doi: #1}\else
  \providecommand{\doi}{doi: \begingroup \urlstyle{rm}\Url}\fi

\bibitem[Ahn et~al.(2022)Ahn, Chen, Wang, and Song]{ahn2022spanning}
Sungsoo Ahn, Binghong Chen, Tianzhe Wang, and Le~Song.
\newblock Spanning tree-based graph generation for molecules.
\newblock In \emph{International Conference on Learning Representations}, 2022.
\newblock URL \url{https://openreview.net/forum?id=w60btE_8T2m}.

\bibitem[Albert \& Barab{\'a}si(2002)Albert and Barab{\'a}si]{albert2002statistical}
R{\'e}ka Albert and Albert-L{\'a}szl{\'o} Barab{\'a}si.
\newblock Statistical mechanics of complex networks.
\newblock \emph{Reviews of modern physics}, 74\penalty0 (1):\penalty0 47, 2002.

\bibitem[Barik et~al.(2020)Barik, Minutoli, Halappanavar, Tallent, and Kalyanaraman]{barik2020vertex}
Reet Barik, Marco Minutoli, Mahantesh Halappanavar, Nathan~R Tallent, and Ananth Kalyanaraman.
\newblock Vertex reordering for real-world graphs and applications: An empirical evaluation.
\newblock In \emph{2020 IEEE International Symposium on Workload Characterization (IISWC)}, pp.\  240--251. IEEE, 2020.

\bibitem[Besta \& Hoefler(2018)Besta and Hoefler]{besta2018survey}
Maciej Besta and Torsten Hoefler.
\newblock Survey and taxonomy of lossless graph compression and space-efficient graph representations.
\newblock \emph{arXiv preprint arXiv:1806.01799}, 2018.

\bibitem[Boldi et~al.(2009)Boldi, Santini, and Vigna]{boldi2009permuting}
Paolo Boldi, Massimo Santini, and Sebastiano Vigna.
\newblock Permuting web and social graphs.
\newblock \emph{Internet Mathematics}, 6\penalty0 (3):\penalty0 257--283, 2009.

\bibitem[Bouritsas et~al.(2021)Bouritsas, Loukas, Karalias, and Bronstein]{bouritsas2021partition}
Giorgos Bouritsas, Andreas Loukas, Nikolaos Karalias, and Michael Bronstein.
\newblock Partition and code: learning how to compress graphs.
\newblock \emph{Advances in Neural Information Processing Systems}, 34:\penalty0 18603--18619, 2021.

\bibitem[Brisaboa et~al.(2009)Brisaboa, Ladra, and Navarro]{brisaboa2009k2}
Nieves~R Brisaboa, Susana Ladra, and Gonzalo Navarro.
\newblock k2-trees for compact web graph representation.
\newblock In \emph{SPIRE}, volume~9, pp.\  18--30. Springer, 2009.

\bibitem[Chen et~al.(2023)Chen, He, Han, and Liu]{chen2023efficient}
Xiaohui Chen, Jiaxing He, Xu~Han, and Li-Ping Liu.
\newblock Efficient and degree-guided graph generation via discrete diffusion modeling.
\newblock \emph{arXiv preprint arXiv:2305.04111}, 2023.

\bibitem[Cuthill \& McKee(1969)Cuthill and McKee]{cuthill1969reducing}
Elizabeth Cuthill and James McKee.
\newblock Reducing the bandwidth of sparse symmetric matrices.
\newblock In \emph{Proceedings of the 1969 24th national conference}, pp.\  157--172, 1969.

\bibitem[Diamant et~al.(2023)Diamant, Tseng, Chuang, Biancalani, and Scalia]{diamant2023improving}
Nathaniel~Lee Diamant, Alex~M Tseng, Kangway~V Chuang, Tommaso Biancalani, and Gabriele Scalia.
\newblock Improving graph generation by restricting graph bandwidth.
\newblock In \emph{International Conference on Machine Learning}, pp.\  7939--7959. PMLR, 2023.

\bibitem[Erd{\H{o}}s et~al.(1960)Erd{\H{o}}s, R{\'e}nyi, et~al.]{erdHos1960evolution}
Paul Erd{\H{o}}s, Alfr{\'e}d R{\'e}nyi, et~al.
\newblock On the evolution of random graphs.
\newblock \emph{Publ. Math. Inst. Hung. Acad. Sci}, 5\penalty0 (1):\penalty0 17--60, 1960.

\bibitem[Goyal et~al.(2020)Goyal, Jain, and Ranu]{goyal2020graphgen}
Nikhil Goyal, Harsh~Vardhan Jain, and Sayan Ranu.
\newblock Graphgen: a scalable approach to domain-agnostic labeled graph generation.
\newblock In \emph{Proceedings of The Web Conference 2020}, pp.\  1253--1263, 2020.

\bibitem[Grover et~al.(2019)Grover, Zweig, and Ermon]{grover2019graphite}
Aditya Grover, Aaron Zweig, and Stefano Ermon.
\newblock Graphite: Iterative generative modeling of graphs.
\newblock In \emph{International conference on machine learning}, pp.\  2434--2444. PMLR, 2019.

\bibitem[Irwin et~al.(2012)Irwin, Sterling, Mysinger, Bolstad, and Coleman]{irwin2012zinc}
John~J Irwin, Teague Sterling, Michael~M Mysinger, Erin~S Bolstad, and Ryan~G Coleman.
\newblock Zinc: a free tool to discover chemistry for biology.
\newblock \emph{Journal of chemical information and modeling}, 52\penalty0 (7):\penalty0 1757--1768, 2012.

\bibitem[Jin et~al.(2018)Jin, Barzilay, and Jaakkola]{jin2018junction}
Wengong Jin, Regina Barzilay, and Tommi Jaakkola.
\newblock Junction tree variational autoencoder for molecular graph generation.
\newblock In \emph{International conference on machine learning}, pp.\  2323--2332. PMLR, 2018.

\bibitem[Jin et~al.(2020)Jin, Barzilay, and Jaakkola]{jin2020hierarchical}
Wengong Jin, Regina Barzilay, and Tommi Jaakkola.
\newblock Hierarchical generation of molecular graphs using structural motifs.
\newblock In \emph{International conference on machine learning}, pp.\  4839--4848. PMLR, 2020.

\bibitem[Jo et~al.(2022)Jo, Lee, and Hwang]{jo2022gdss}
Jaehyeong Jo, Seul Lee, and Sung~Ju Hwang.
\newblock Score-based generative modeling of graphs via the system of stochastic differential equations.
\newblock In \emph{International Conference on Machine Learning}, pp.\  10362--10383. PMLR, 2022.

\bibitem[Kong et~al.(2023)Kong, Cui, Sun, Zhuang, Prakash, and Zhang]{kong2023autoregressive}
Lingkai Kong, Jiaming Cui, Haotian Sun, Yuchen Zhuang, B.~Aditya Prakash, and Chao Zhang.
\newblock Autoregressive diffusion model for graph generation, 2023.
\newblock URL \url{https://openreview.net/forum?id=98J48HZXxd5}.

\bibitem[Krenn et~al.(2019)Krenn, H{\"a}se, Nigam, Friederich, and Aspuru-Guzik]{krenn2019selfies}
Mario Krenn, Florian H{\"a}se, A~Nigam, Pascal Friederich, and Al{\'a}n Aspuru-Guzik.
\newblock {SELFIES}: a robust representation of semantically constrained graphs with an example application in chemistry.
\newblock \emph{arXiv preprint arXiv:1905.13741}, 2019.

\bibitem[Larsson \& Moffat(2000)Larsson and Moffat]{larsson2000off}
N~Jesper Larsson and Alistair Moffat.
\newblock Off-line dictionary-based compression.
\newblock \emph{Proceedings of the IEEE}, 88\penalty0 (11):\penalty0 1722--1732, 2000.

\bibitem[Li et~al.(2018)Li, Vinyals, Dyer, Pascanu, and Battaglia]{li2018learning}
Yujia Li, Oriol Vinyals, Chris Dyer, Razvan Pascanu, and Peter Battaglia.
\newblock Learning deep generative models of graphs.
\newblock \emph{arXiv preprint arXiv:1803.03324}, 2018.

\bibitem[Liao et~al.(2019)Liao, Li, Song, Wang, Hamilton, Duvenaud, Urtasun, and Zemel]{liao2019efficient}
Renjie Liao, Yujia Li, Yang Song, Shenlong Wang, Will Hamilton, David~K Duvenaud, Raquel Urtasun, and Richard Zemel.
\newblock Efficient graph generation with graph recurrent attention networks.
\newblock \emph{Advances in neural information processing systems}, 32, 2019.

\bibitem[Liu et~al.(2019)Liu, Kumar, Ba, Kiros, and Swersky]{liu2019gnf}
Jenny Liu, Aviral Kumar, Jimmy Ba, Jamie Kiros, and Kevin Swersky.
\newblock Graph normalizing flows.
\newblock \emph{Advances in Neural Information Processing Systems}, 32, 2019.

\bibitem[Liu et~al.(2021)Liu, Yan, Oztekin, and Ji]{liu2021graphebm}
Meng Liu, Keqiang Yan, Bora Oztekin, and Shuiwang Ji.
\newblock Graphebm: Molecular graph generation with energy-based models.
\newblock \emph{arXiv preprint arXiv:2102.00546}, 2021.

\bibitem[Luo et~al.(2022)Luo, Mo, and Pan]{luo2022fast}
Tianze Luo, Zhanfeng Mo, and Sinno~Jialin Pan.
\newblock Fast graph generative model via spectral diffusion.
\newblock \emph{arXiv preprint arXiv:2211.08892}, 2022.

\bibitem[Luo et~al.(2021)Luo, Yan, and Ji]{luo2021graphdf}
Youzhi Luo, Keqiang Yan, and Shuiwang Ji.
\newblock Graphdf: A discrete flow model for molecular graph generation.
\newblock In \emph{International Conference on Machine Learning}, pp.\  7192--7203. PMLR, 2021.

\bibitem[Madhawa et~al.(2019)Madhawa, Ishiguro, Nakago, and Abe]{madhawa2019graphnvp}
Kaushalya Madhawa, Katushiko Ishiguro, Kosuke Nakago, and Motoki Abe.
\newblock Graphnvp: An invertible flow model for generating molecular graphs.
\newblock \emph{arXiv preprint arXiv:1905.11600}, 2019.

\bibitem[Martinkus et~al.(2022)Martinkus, Loukas, Perraudin, and Wattenhofer]{martinkus2022spectre}
Karolis Martinkus, Andreas Loukas, Nathana{\"e}l Perraudin, and Roger Wattenhofer.
\newblock Spectre: Spectral conditioning helps to overcome the expressivity limits of one-shot graph generators.
\newblock In \emph{International Conference on Machine Learning}, pp.\  15159--15179. PMLR, 2022.

\bibitem[Maziarka et~al.(2020)Maziarka, Pocha, Kaczmarczyk, Rataj, Danel, and Warcho{\l}]{maziarka2020mol}
{\L}ukasz Maziarka, Agnieszka Pocha, Jan Kaczmarczyk, Krzysztof Rataj, Tomasz Danel, and Micha{\l} Warcho{\l}.
\newblock Mol-cyclegan: a generative model for molecular optimization.
\newblock \emph{Journal of Cheminformatics}, 12\penalty0 (1):\penalty0 1--18, 2020.

\bibitem[Mueller(2004)]{mueller2004sparse}
Chris Mueller.
\newblock Sparse matrix reordering algorithms for cluster identification.
\newblock \emph{Machune Learning in Bioinformatics}, 2004.

\bibitem[Niu et~al.(2020)Niu, Song, Song, Zhao, Grover, and Ermon]{niu2020edpgnn}
Chenhao Niu, Yang Song, Jiaming Song, Shengjia Zhao, Aditya Grover, and Stefano Ermon.
\newblock Permutation invariant graph generation via score-based generative modeling.
\newblock In \emph{International Conference on Artificial Intelligence and Statistics}, pp.\  4474--4484. PMLR, 2020.

\bibitem[Paszke et~al.(2019)Paszke, Gross, Massa, Lerer, Bradbury, Chanan, Killeen, Lin, Gimelshein, Antiga, et~al.]{paszke2019pytorch}
Adam Paszke, Sam Gross, Francisco Massa, Adam Lerer, James Bradbury, Gregory Chanan, Trevor Killeen, Zeming Lin, Natalia Gimelshein, Luca Antiga, et~al.
\newblock Pytorch: An imperative style, high-performance deep learning library.
\newblock \emph{Advances in neural information processing systems}, 32, 2019.

\bibitem[Raghavan \& Garcia-Molina(2003)Raghavan and Garcia-Molina]{raghavan2003representing}
Sriram Raghavan and Hector Garcia-Molina.
\newblock Representing web graphs.
\newblock In \emph{Proceedings 19th International Conference on Data Engineering (Cat. No. 03CH37405)}, pp.\  405--416. IEEE, 2003.

\bibitem[Ramakrishnan et~al.(2014)Ramakrishnan, Dral, Rupp, and Von~Lilienfeld]{ramakrishnan2014quantum}
Raghunathan Ramakrishnan, Pavlo~O Dral, Matthias Rupp, and O~Anatole Von~Lilienfeld.
\newblock Quantum chemistry structures and properties of 134 kilo molecules.
\newblock \emph{Scientific data}, 1\penalty0 (1):\penalty0 1--7, 2014.

\bibitem[Schomburg et~al.(2004)Schomburg, Chang, Ebeling, Gremse, Heldt, Huhn, and Schomburg]{schomburg2004brenda}
Ida Schomburg, Antje Chang, Christian Ebeling, Marion Gremse, Christian Heldt, Gregor Huhn, and Dietmar Schomburg.
\newblock Brenda, the enzyme database: updates and major new developments.
\newblock \emph{Nucleic acids research}, 32\penalty0 (suppl\_1):\penalty0 D431--D433, 2004.

\bibitem[Segler et~al.(2018)Segler, Kogej, Tyrchan, and Waller]{segler2018generating}
Marwin~HS Segler, Thierry Kogej, Christian Tyrchan, and Mark~P Waller.
\newblock Generating focused molecule libraries for drug discovery with recurrent neural networks.
\newblock \emph{ACS central science}, 4\penalty0 (1):\penalty0 120--131, 2018.

\bibitem[Shaw et~al.(2018)Shaw, Uszkoreit, and Vaswani]{shaw2018self}
Peter Shaw, Jakob Uszkoreit, and Ashish Vaswani.
\newblock Self-attention with relative position representations.
\newblock \emph{arXiv preprint arXiv:1803.02155}, 2018.

\bibitem[Shi et~al.(2020)Shi, Xu, Zhu, Zhang, Zhang, and Tang]{shi2020graphaf}
Chence Shi, Minkai Xu, Zhaocheng Zhu, Weinan Zhang, Ming Zhang, and Jian Tang.
\newblock Graphaf: a flow-based autoregressive model for molecular graph generation.
\newblock \emph{arXiv preprint arXiv:2001.09382}, 2020.

\bibitem[Simonovsky \& Komodakis(2018)Simonovsky and Komodakis]{simonovsky2018graphvae}
Martin Simonovsky and Nikos Komodakis.
\newblock Graphvae: Towards generation of small graphs using variational autoencoders.
\newblock In \emph{Artificial Neural Networks and Machine Learning--ICANN 2018: 27th International Conference on Artificial Neural Networks, Rhodes, Greece, October 4-7, 2018, Proceedings, Part I 27}, pp.\  412--422. Springer, 2018.

\bibitem[Vaswani et~al.(2017)Vaswani, Shazeer, Parmar, Uszkoreit, Jones, Gomez, Kaiser, and Polosukhin]{vaswani2017attention}
Ashish Vaswani, Noam Shazeer, Niki Parmar, Jakob Uszkoreit, Llion Jones, Aidan~N Gomez, {\L}ukasz Kaiser, and Illia Polosukhin.
\newblock Attention is all you need.
\newblock \emph{Advances in neural information processing systems}, 30, 2017.

\bibitem[Vignac et~al.(2022)Vignac, Krawczuk, Siraudin, Wang, Cevher, and Frossard]{vignac2022digress}
Clement Vignac, Igor Krawczuk, Antoine Siraudin, Bohan Wang, Volkan Cevher, and Pascal Frossard.
\newblock Digress: Discrete denoising diffusion for graph generation.
\newblock \emph{arXiv preprint arXiv:2209.14734}, 2022.

\bibitem[Yang et~al.(2021)Yang, Hwang, Lee, Ryu, and Hwang]{yang2021hit}
Soojung Yang, Doyeong Hwang, Seul Lee, Seongok Ryu, and Sung~Ju Hwang.
\newblock Hit and lead discovery with explorative rl and fragment-based molecule generation.
\newblock \emph{Advances in Neural Information Processing Systems}, 34:\penalty0 7924--7936, 2021.

\bibitem[Ying et~al.(2018)Ying, You, Morris, Ren, Hamilton, and Leskovec]{ying2018hierarchical}
Zhitao Ying, Jiaxuan You, Christopher Morris, Xiang Ren, Will Hamilton, and Jure Leskovec.
\newblock Hierarchical graph representation learning with differentiable pooling.
\newblock \emph{Advances in neural information processing systems}, 31, 2018.

\bibitem[You et~al.(2018)You, Ying, Ren, Hamilton, and Leskovec]{you2018graphrnn}
Jiaxuan You, Rex Ying, Xiang Ren, William Hamilton, and Jure Leskovec.
\newblock Graphrnn: Generating realistic graphs with deep auto-regressive models.
\newblock In \emph{International conference on machine learning}, pp.\  5708--5717. PMLR, 2018.

\bibitem[Yu et~al.(2020)Yu, Sun, Solvang, and Zhao]{yu2020reverse}
Hao Yu, Xu~Sun, Wei~Deng Solvang, and Xu~Zhao.
\newblock Reverse logistics network design for effective management of medical waste in epidemic outbreaks: Insights from the coronavirus disease 2019 (covid-19) outbreak in wuhan (china).
\newblock \emph{International journal of environmental research and public health}, 17\penalty0 (5):\penalty0 1770, 2020.

\bibitem[Zang \& Wang(2020)Zang and Wang]{zang2020moflow}
Chengxi Zang and Fei Wang.
\newblock Moflow: an invertible flow model for generating molecular graphs.
\newblock In \emph{Proceedings of the 26th ACM SIGKDD International Conference on Knowledge Discovery \& Data Mining}, pp.\  617--626, 2020.

\end{thebibliography}
\bibliographystyle{iclr2024_conference}

\newpage
\appendix
\section{Construction of a \tree from the graph}\label{appx:graphtok2}
%\setlength{\intextsep}{0pt}%
%\setlength{\columnsep}{11pt}%
%\begin{wrapfigure}{R}{0.6\textwidth}
%\begin{minipage}[t]{\linewidth}
\begin{algorithm}[h]
\caption{\tree construction}
\label{algo:k2tree}
\begin{algorithmic}[1]
\Statex \textbf{Input}:Adjacency matrix $A$ and partitioning~factor $K$.
%\Statex \textbf{Output}: \tree $T$
\State Initialize the tree $\mathcal{T} \gets (\mathcal{V}, \mathcal{E})$ with $\mathcal{V}= \emptyset, \mathcal{E}= \emptyset$. \Comment{\tree.}
\State Initialize an empty queue $Q$. \Comment{Candidates to be expanded into child nodes.}
\State Set $\mathcal{V} \gets \mathcal{V} \cup \{r\}$, $x_r \gets 1$ and let $A^{(r)} \leftarrow A$. Insert $r$ into the queue $Q$. \Comment{Add root node $r$.}
\While{$Q \neq \emptyset$}
    \State Pop $u$ from $\mathcal{Q}$.
    % \If{$A^{(u)}$ is not filled with zeros} \Comment{Expand child nodes.}
    \If{$x_u = 0$} \Comment{Condition for not expanding the node $u$.}
    \State Go to line 4.
    \EndIf
    \State Update $s \gets \operatorname{dim}(A^{(u)}) / K$
    \For{$i=1, \ldots, K$} \Comment{Row-wise indices.}
        \For{$j=1, \ldots, K$} \Comment{Column-wise indices.}
            \State Set $B_{i, j} \gets A^{(u)}[(i-1)s:is, (j-1)s:js]$.
            \Statex\Comment{Operation to obtain $s\times s$ submatrix $B_{i,j}$ of $A^{(u)}$.}
            \State If $B_{i, j}$ is filled with zeros, set $x_{v} \gets 1$.
            Otherwise, set $x_v \gets 0$. 
            \Statex\Comment{Update tree-node attribute.}
            \State If $\operatorname{dim}(v_{i,j}) > 1$, update $\mathcal{Q} \gets v_{i,j}$.
        \EndFor
    \EndFor
    % \State Set $\mathcal{V} \gets \mathcal{V} \cup \{N_{\mathcal{V}}+1, \ldots, N_{\mathcal{V}}+K^{2}\}$.
    \State Set $\mathcal{V} \gets \mathcal{V} \cup \{v_{1,1}, \ldots, v_{K,K}\}$. \Comment{Update tree nodes.}
    % \State Set $\mathcal{E} \gets \mathcal{E} \cup \{(i, N_{\mathcal{V}}+1), \ldots, (i, N_{\mathcal{V}}+K^{2})\}$.
    \State Set $\mathcal{E} \gets \mathcal{E} \cup \{(u, v_{1,1}), \ldots, (u, v_{K,K})\}$. \Comment{Update tree edges.}
    % \State Set $N_{\mathcal{V}}\gets N_{\mathcal{V}}+K^{2}$.
\EndWhile
% \Return $(\mathcal{T}, \mathcal{X})$
\Statex \textbf{Output}: \tree $(\mathcal{T}, \mathcal{X})$ where $\mathcal{X} = \{x_{u}: u\in \mathcal{V}\}$.
\end{algorithmic}
\end{algorithm}
%\end{minipage}
%\end{wrapfigure}

In this section, we explain our algorithm to construct a \tree $(\mathcal{T}, \mathcal{X})$ from a given graph $G=A$ where $G$ is a symmetric non-featured graph and $A$ is an adjacency matrix. Note that he $K^{2}$-ary tree $\mathcal{T} = (\mathcal{V}, \mathcal{E})$ is associated with binary node attributes $\mathcal{X} = \{x_{u}: u\in\mathcal{V}\}$. In addition, let $\operatorname{dim}(A)$ to denote the number of rows(or columns) $n$ of the square matrix $A \in \{0, 1\}^{n \times n}$. We describe the full procedure in \cref{algo:k2tree}. \textcolor{black}{Note that the time complexity of the procedure is $O(N^2)$ \citep{brisaboa2009k2}, where $N$ denotes the number of nodes in the graph $G$.}

\newpage
\section{Constructing a graph from the \tree}\label{appx:k2tograph}
\begin{algorithm}[h]
\caption{Graph $G$ construction}
\label{algo:graph}
\begin{algorithmic}
\Statex \textbf{Input}: \tree $(\mathcal{T}, \mathcal{X})$ and partitioning factor $K$.
\Statex Set $m \gets K^{D_{\mathcal{T}}}$. \Comment{Full adjacency matrix size.}
\Statex Initialize $A \in \{0, 1\}^{m \times m}$ with zeros. 
\For{$u \in \mathcal{L}$} \Comment{For each leaf node with $x_u=1$.}
    \State $\operatorname{pos}(u)=((i_{v_1}, j_{v_1}), \ldots, (i_{v_L}, j_{v_L}))$. \Comment{Position of node $u$.}
    \State ${(p_u,q_u)= (\sum_{\ell=1}^{L}K^{L-\ell}(i_{v_{\ell}}-1)+1, \sum_{\ell=1}^{L}K^{L-\ell}(j_{v_{\ell}}-1)+1)}$. \Comment{Location of node $u$.}
    \State Set $A_{p_u, q_u} \gets 1$.
\EndFor

\Statex \textbf{Output}: Adjacency matrix $A$.
\end{algorithmic}
\end{algorithm}  
We next describe the algorithm to generate a graph $G=A$ given the \tree $(\mathcal{T}, \mathcal{X})$ with tree depth $D_{\mathcal{T}}$. Let $\mathcal{L} \subset \mathcal{V}$ be the set of leaf nodes in \tree with node attributes $1$. Note that we represent the tree-node position of $u \in \mathcal{V}$ as $\operatorname{pos}(u)=((i_{v_1}, j_{v_1}), \ldots, (i_{v_L}, j_{v_L}))$ based on a downward path $v_{0}, v_{1}, \ldots, v_{L}$ from the root node $r=v_{0}$ to the tree-node $u=v_{L}$. In addition, the location of corresponding submatrix $A^{(u)}$ is denoted as ${(p_u,q_u)= (\sum_{\ell=1}^{L}K^{L-\ell}(i_{v_{\ell}}-1)+1, \sum_{\ell=1}^{L}K^{L-\ell}(j_{v_{\ell}}-1)+1)}$ in as described in \cref{subsec:seqrep}. We describe the full procedure as in \cref{algo:graph}.  \textcolor{black}{Note that the time complexity of the procedure is $O(N^2)$, where $N$ denotes the number of nodes in the graph $G$, since it requires querying for each element in the adjacency matrix.}

\newpage
\section{Generalizing \tree to Attributed Graphs}\label{appx:attr}
\begin{algorithm}[h]
\caption{Featured \tree construction}
\label{algo:k2tree_feature}
\begin{algorithmic}[1]
\Statex \textbf{Input}: Modified adjacency matrix $A$ and partitioning~factor $K$.
%\Statex \textbf{Output}: \tree $T$
\State Initialize the tree $\mathcal{T} \gets (\mathcal{V}, \mathcal{E})$ with $\mathcal{V}= \emptyset, \mathcal{E}= \emptyset$. \Comment{Featured \tree.}
\State Initialize an empty queue $Q$. \Comment{Candidates to be expanded into child nodes.}
\State Set $\mathcal{V} \gets \mathcal{V} \cup \{r\}$, $x_r \gets 1$ and let $A^{(r)} \leftarrow A$. Insert $r$ into the queue $Q$. \Comment{Add root node $r$.}
\While{$Q \neq \emptyset$}
    \State Pop $u$ from $\mathcal{Q}$.
    \If{$x_u = 0$} \Comment{Condition for not expanding the node $u$.}
    \State Go to line 4.
    \EndIf
    \State Update $s \gets \operatorname{dim}(A^{(u)}) / K$
    \For{$i=1, \ldots, K$} \Comment{Row-wise indices.}
        \For{$j=1, \ldots, K$} \Comment{Column-wise indices.}
            \State Set $B_{i, j} \gets A^{(u)}[(i-1)s:is, (j-1)s:js]$.
            \Statex\Comment{Operation to obtain $s\times s$ submatrix $B_{i,j}$ of $A^{(u)}$.}
            \If{$B_{i, j}$ is filled with zeros} \Comment{Update tree-node attribute.}
            \State Set $x_v \gets 0$. 
            \ElsIf {$|B_{i,j}| > 1$} \Comment{Non-leaf tree-nodes with attribute 1.}
            \State Set $x_v \gets 1$.
            \Else \Comment{Leaf tree-nodes with node features and edge features.}
            \State Set $x_v \gets B_{i,j}$. \Comment{We treat $1\times 1$ matrix $B_{i, j}$ as a scalar.}
            \EndIf
            \If{$\operatorname{dim}(B_{i,j}) > 1$} $\mathcal{Q} \gets v_{i,j}$.
            \EndIf
        \EndFor
    \EndFor
    \State Set $\mathcal{V} \gets \mathcal{V} \cup \{v_{1,1}, \ldots, v_{K,K}\}$. \Comment{Update tree nodes.}
    \State Set $\mathcal{E} \gets \mathcal{E} \cup \{(u, v_{1,1}), \ldots, (u, v_{K,K})\}$. \Comment{Update tree edges.}
    % \State Set $N_{\mathcal{V}}\gets N_{\mathcal{V}}+K^{2}$.
\EndWhile
% \Return $(\mathcal{T}, \mathcal{X})$
\Statex \textbf{Output}: Featured \tree $(\mathcal{T}, \mathcal{X})$ where $\mathcal{X} = \{x_{u}: u\in \mathcal{V}\}$.
\end{algorithmic}
\end{algorithm}
%\end{minipage}
%\end{wrapfigure}
\begin{algorithm}[h]
\caption{Featured graph $G$ construction}
\label{algo:graph_featured}
\begin{algorithmic}[1]
\State \textbf{Input}: Featured \tree $(\mathcal{T}, \mathcal{X})$ and partitioning factor $K$.
\State $m \gets K^{D_{\mathcal{T}}}$ \Comment{Full adjacency matrix size.}
\State Initialize $A \in \{0, 1\}^{m \times m}$ with zeros. 
\For{$u \in \mathcal{L}$} \Comment{For each leaf node with $x_u \neq 0$.}
    \State $\operatorname{pos}(u)=((i_{v_1}, j_{v_1}), \ldots, (i_{v_L}, j_{v_L}))$. \Comment{Position of node $u$.}
    \State ${(p_u,q_u)= (\sum_{\ell=1}^{L}K^{L-\ell}(i_{v_{\ell}}-1)+1, \sum_{\ell=1}^{L}K^{L-\ell}(j_{v_{\ell}}-1)+1)}$. \Comment{Location of node $u$.}
    \State Set $A_{p_u, q_u} \gets x_u$.
\EndFor

\State \textbf{Output}: Modified adjacency matrix $A$.
\end{algorithmic}
\end{algorithm}  

In this section, we describe a detailed process to construct a \tree for featured graphs with node features and edge features (e.g., molecular graphs), which is described briefly in \cref{subsec:seqrep}. We modify the original adjacency matrix by incorporating categorical features into each element, thereby enabling the derivation of the featured \tree from the modified adjacency matrix.

\textbf{Edge features.} Integrating edge features into the adjacency matrix is straightforward. It can be accomplished by simply replacing the ones with the appropriate categorical edge features.

\textbf{Node features.} Integrating node features into the adjacency matrix is more complex than that of edge features since the adjacency matrix only describes the connectivity between node pairs. To address this issue, we assume that all graph nodes possess self-loops, which leads to filling ones to the diagonal elements. Then we replace ones on the diagonal with categorical node features that correspond to the respective node positions.

Let $x_u \in \mathcal{X}$ be the non-binary tree-node attributes that include node features and edge features and $\mathcal{L}$ be the set of leaf nodes in \tree with non-zero node attributes. Then we can construct a featured \tree with a modified adjacency matrix and construct a graph $G$ from the featured \tree as described in \cref{algo:k2tree_feature} and \cref{algo:graph_featured}, respectively.

\clearpage
\newpage

\section{Experimental Details}\label{appx:exp}
In this section, we provide the details of the experiments. Note that we chose $k=2$ in all experiments and provide additional experimental results for $k=3$ in \cref{appx:additional_result}.

\subsection{Generic graph generation}
\begin{table}[h]
\vspace{-.1in}
  \caption{\textbf{Hyperparameters of \Algname in generic graph generation.}}
  \label{tab:appx_hyper}
  \centering
  \scalebox{0.8}{
    \begin{tabular}{llcccc}
    % \toprule & Ego-small & Community-small & Enzymes & Grid & Planar & SBM & Proteins \\
    \toprule 
    & Hyperparameter & Community-small & Planar & Enzymes & Grid \\
    \midrule
    \multirow{4}{*}{Transformer}
    & Dim. of feed-forward network & 512 & 512 & 512 & 512  \\
    & Transformer dropout rate & 0.1 & 0 & 0.1 & 0.1  \\
    & \# of attention heads & 8 & 8 & 8 & 8 \\
    & \# of layers & 3 & 3 & 3 & 3 \\
    \midrule
    \multirow{6}{*}{Train}
    & Batch size & 128  & 32 & 32 & 8 \\
    & \# of epochs & 500 & 500 & 500 & 500\\
    & Dim. of token embedding & 512 & 512 & 512 & 512 \\
    & Gradient clipping norm & 1 & 1 & 1 & 1 \\
    & Input dropout rate & 0 & 0 & 0 & 0 \\
    & Learning rate & $1 \times 10^{-3}$ & $1 \times 10^{-3}$ & $2 \times 10^{-4}$ & $5 \times 10^{-4}$ \\
    \bottomrule
  \end{tabular}
  }
\end{table}
We used the same split with GDSS \citep{jo2022gdss} for Community-small, Enzymes, and Grid datasets. Otherwise, we used the same split with SPECTRE \citep{luo2022fast} for the Planar dataset. We fix $k=2$ and perform the hyperparameter search to choose the best learning rate in $\{0.0001, 0.0002, 0.0005, 0.001\}$ and the best dropout rate in $\{0, 0.1\}$. We select the model with the best MMD with the lowest average of three graph statistics: degree, clustering coefficient, and orbit count. Finally, we provide the hyperparameters used in the experiment in \cref{tab:appx_hyper}. 

\subsection{Molecular graph generation}
\begin{table}[h]
\vspace{-.1in}
  \caption{\textbf{Statstics of molecular datasets: QM9 and ZINC250k}.}
  \label{tab:mol_stat}
  \centering
  \scalebox{0.8}{
    \begin{tabular}{lcccc}
    % \toprule & Ego-small & Community-small & Enzymes & Grid & Planar & SBM & Proteins \\
    \toprule 
    Dataset & \# of graphs & \# of nodes & \# of node types & \# of edge types \\
    \midrule
    QM9  & 133,885 & $1 \leq |V| \leq 9$ & 4 & 3 \\
    ZINC250k & 249,455 & $6 \leq |V| \leq 38$  & 9 & 3 \\
    \bottomrule
  \end{tabular}
  }
\end{table}
\begin{table}[h]
\vspace{-.1in}
  \caption{\textbf{Hyperparameters of \Algname in molecular graph generation.}}
  \label{tab:appx_hyper}
  \centering
  \scalebox{0.8}{
    \begin{tabular}{llcc}
    % \toprule & Ego-small & Community-small & Enzymes & Grid & Planar & SBM & Proteins \\
    \toprule 
    & Hyperparameter & QM9 & ZINC250k \\
    \midrule
    \multirow{4}{*}{Transformer}
    & Dim. of feedforward network & 512 & 512 \\
    & Transformer dropout rate & 0.1 & 0.1 \\
    & \# of attention heads & 8 & 8 \\
    & \# of layers & 2 & 3 \\
    \midrule
    \multirow{6}{*}{Train}
    & Batch size & 1024 & 256 \\
    & \# of epochs & 500 & 500 \\
    & Dim. of token embedding & 512 & 512 \\
    & Gradient clipping norm & 1 & 1 \\
    & Input dropout rate & 0.5 & 0 \\
    & Learning rate & $5 \times 10^{-4}$ & $5 \times 10^{-4}$ \\
    \bottomrule
  \end{tabular}
  }
%\end{table}
\end{table}
The statistics of training molecular graphs (i.e., QM9 and ZINC250k datasets) are summarized in \cref{tab:mol_stat} and we used the same split with GDSS \citep{jo2022gdss} for a fair evaluation. We fix $k=2$ and perform the hyperparameter search to choose the best number of layers in $\{2, 3\}$ and select the model with the best validity. In addition, we provide the hyperparameters used in the experiment in \cref{tab:appx_hyper}.

\newpage
\clearpage
\section{Implementation Details}\label{appx:impl}
% preprocessing / digress epoch / model architecture
\subsection{Computing resources}
We used PyTorch \citep{paszke2019pytorch} to implement \Algname and train the Transformer \citep{vaswani2017attention} models on \textcolor{black}{a single} GeForce RTX 3090 GPU.

\subsection{Model architecture}
\begin{wrapfigure}{r}{0.45\textwidth}
    \centering
    \vspace{-0.1in}
    \includegraphics[width=\linewidth]{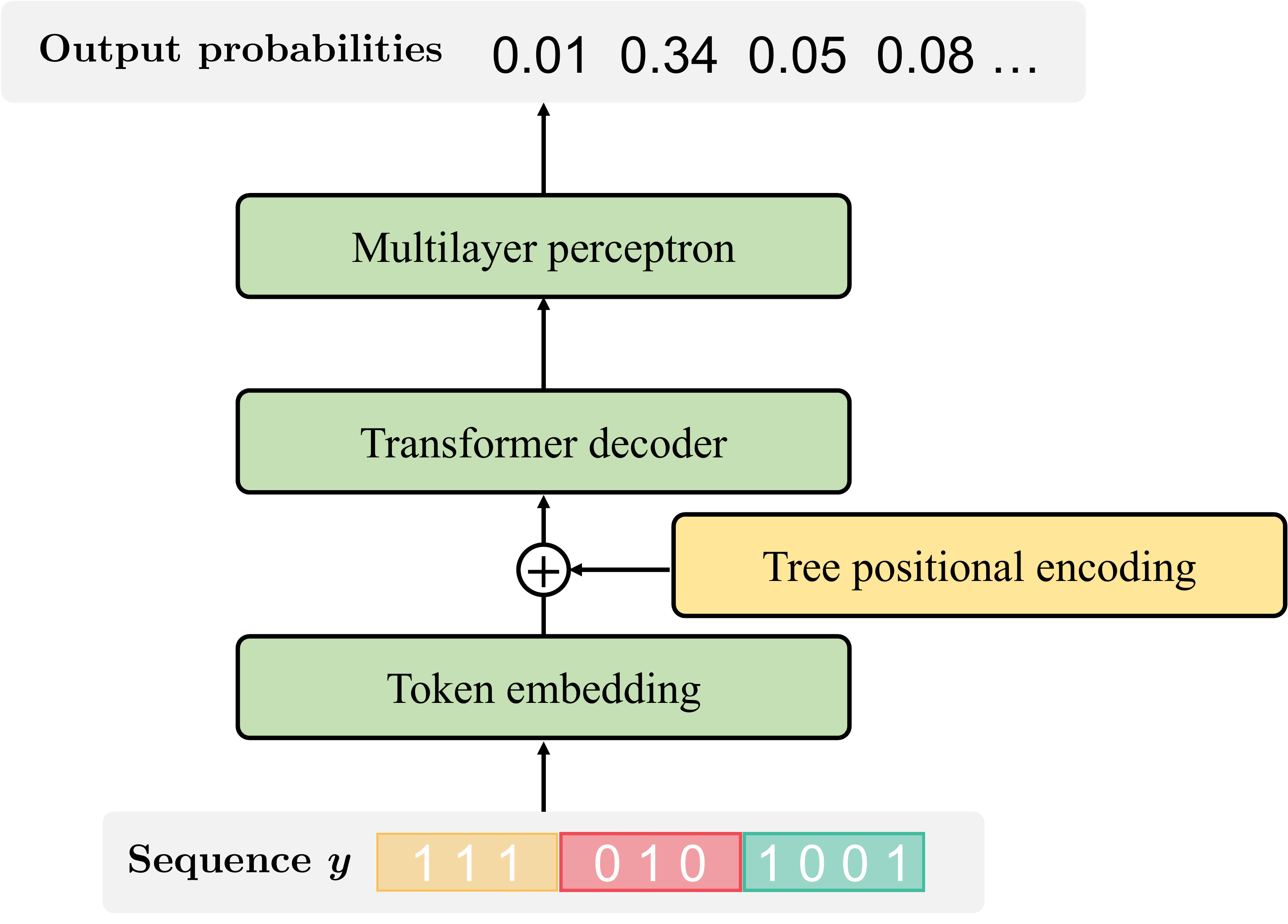}
    \caption{\textbf{The architecture of \Algname.} }
    \label{fig:appx_arch}
    \vspace{-.3in}
\end{wrapfigure}

We describe the architecture of the proposed transformer generator of \Algname in \cref{fig:appx_arch}. The generator takes a sequential representation of \tree as input and generates the output probability of each token as described in \cref{subsec:seqgen}. The model consists of a token embedding layer, transformer encoder(s), and multilayer perceptron layer with tree positional encoding.  

\subsection{Details for baseline implementation}
% DiGress hyperparameter search?
\textbf{Generic graph generation.} The baseline results from prior works are as follows. Results for GraphVAE \citep{simonovsky2018graphvae}, GraphRNN \citep{you2018graphrnn}, GNF \citep{liu2019gnf}, EDP-GNN \citep{niu2020edpgnn}, GraphAF \citep{shi2020graphaf}, GraphDF \citep{luo2021graphdf}, and GDSS \citep{jo2022gdss} are obtained from GDSS, while the results for GRAN \citep{liao2019efficient}, SPECTRE \citep{martinkus2022spectre}, and GDSM \citep{luo2022fast} are derived from their respective paper. Additionally, we reproduced DiGress \citep{vignac2022digress} and GraphGen \citep{goyal2020graphgen} using their open-source codes. We used original hyperparameters when the original work provided them. DiGress takes more than three days for the Planar, Enzymes, and Grid datasets, so we report the results from fewer epochs after convergence.

\textbf{Molecular graph generation.} The baseline results from prior works are as follows. The results for EDP-GNN \citep{niu2020edpgnn}, MoFlow \citep{zang2020moflow}, GraphAF \citep{shi2020graphaf}, GraphDF \citep{luo2021graphdf}, GraphEBM \citep{liu2021graphebm}, and GDSS \citep{jo2022gdss} are from GDSS, and the GDSM \citep{luo2022fast} result is extracted from the corresponding paper.  Moreover, we reproduced DiGress \citep{vignac2022digress} using their open-source codes.

\subsection{Details for the implementation}

\textcolor{black}{We adapted node ordering code from \citep{diamant2023improving}, evaluation scheme from \citep{jo2022gdss, martinkus2022spectre}, and NSPDK computation from \citep{goyal2020graphgen}.}

\newpage
\section{Generated samples}\label{appx:samples}
In this section, we provide the visualizations of the generated graphs for generic and molecular graph generation.

\subsection{Generic graph generation}
\begin{figure}[h]
    \centering
    \includegraphics[scale=0.25]{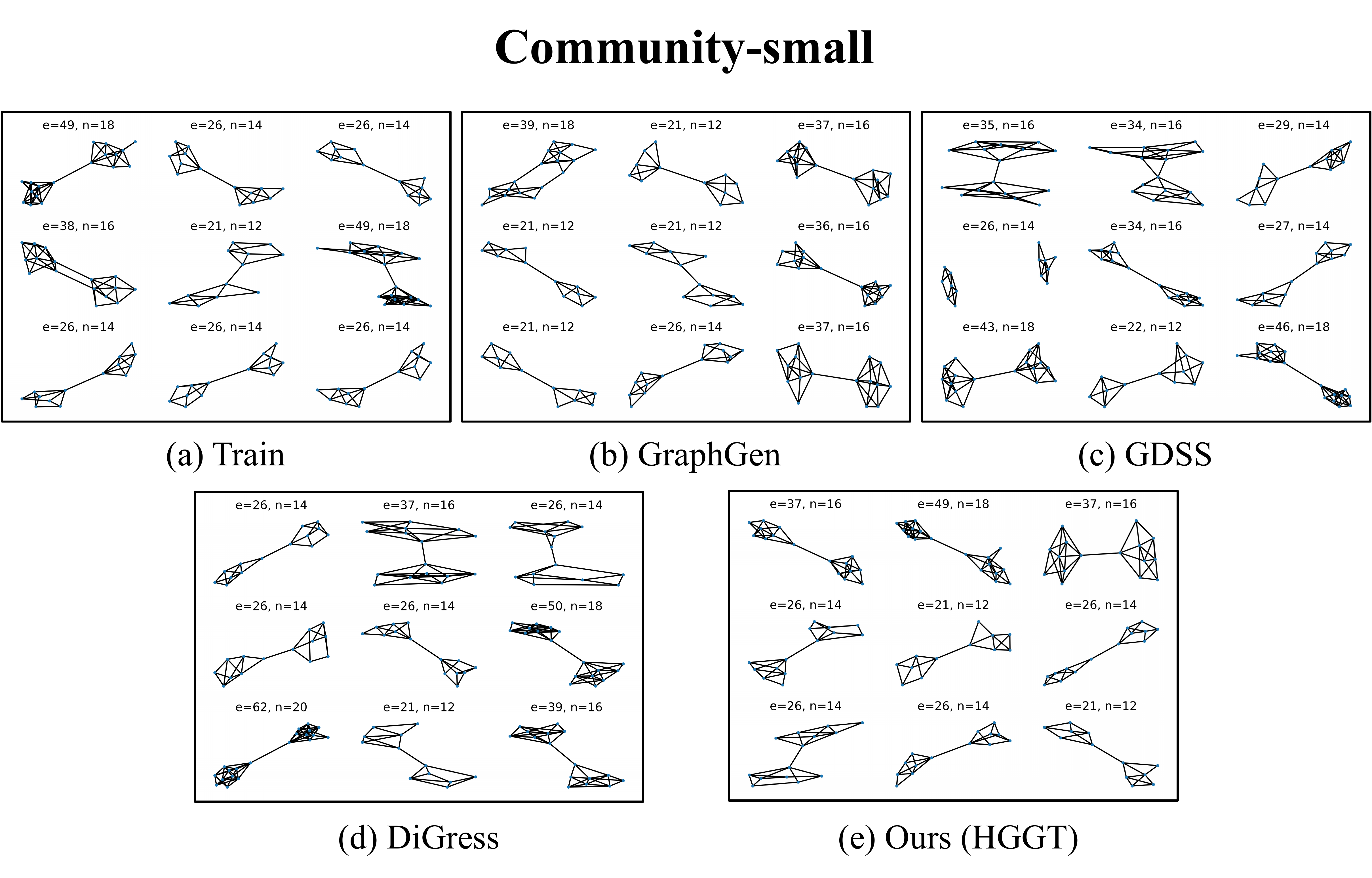}
    \vspace{-0.1in}
    \caption{\small \textbf{Visualization of the graphs from the Community-small dataset and the generated graphs.}}
    \label{fig:community_small}
\end{figure}

\begin{figure}[h]
    \centering
    \includegraphics[scale=0.25]{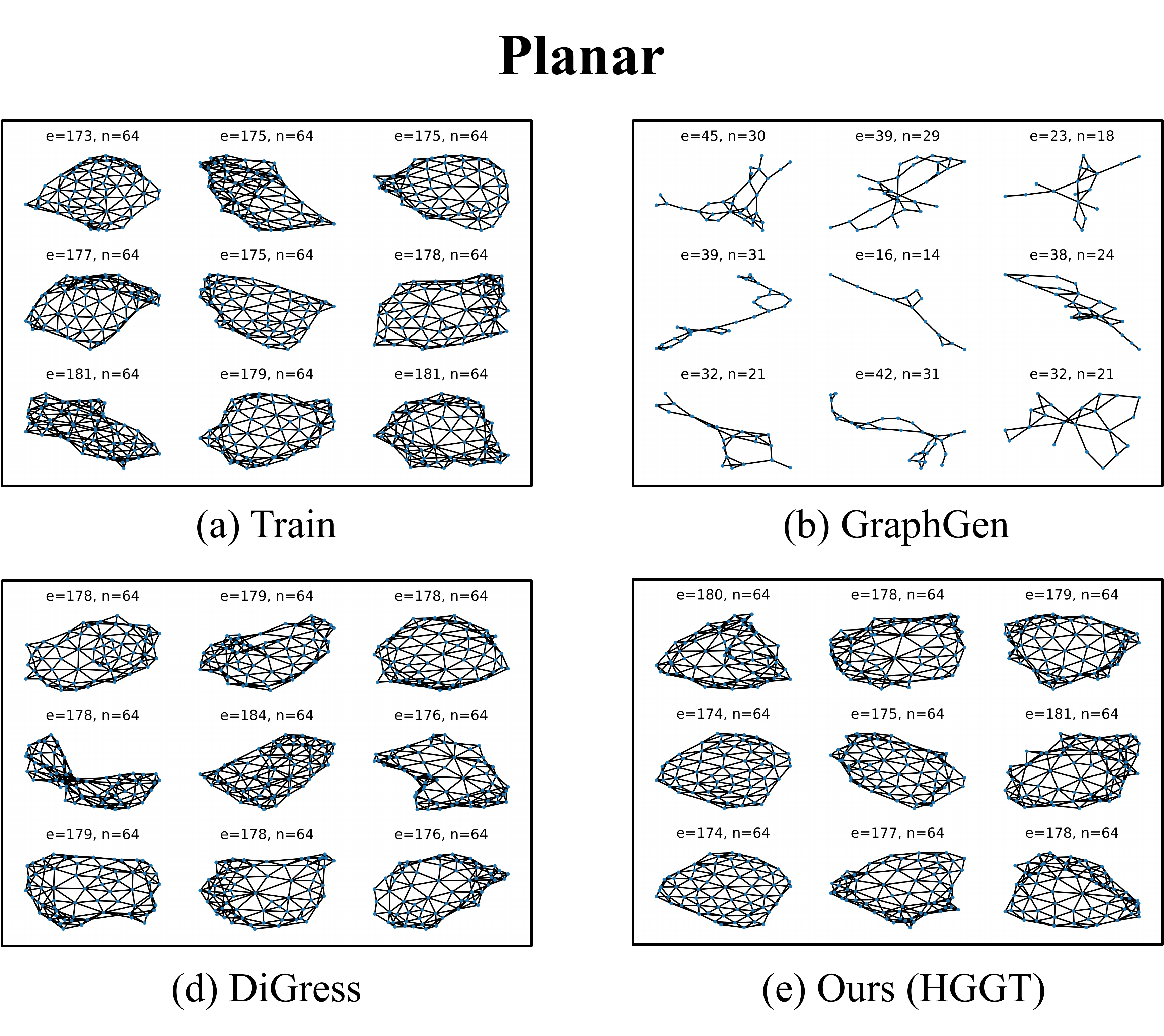}
    \vspace{-0.1in}
    \caption{\small \textbf{Visualization of the graphs from the Planar dataset and the generated graphs.}}
    \label{fig:planar}
\end{figure}

\begin{figure}[h]
    \centering
    \includegraphics[scale=0.25]{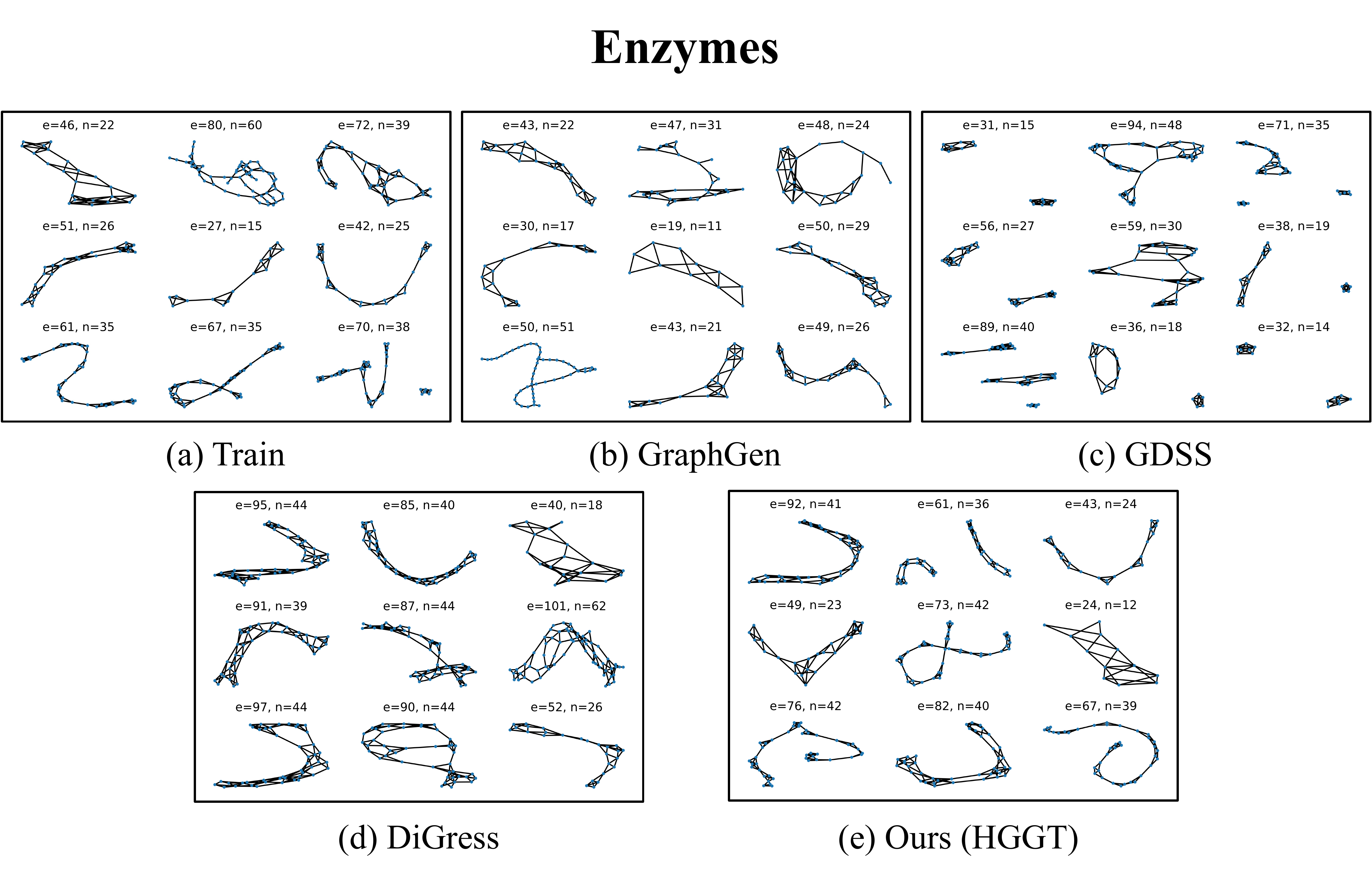}
    \vspace{-0.1in}
    \caption{\small \textbf{Visualization of the graphs from the enzymes dataset and the generated graphs.}}
    \label{fig:enzymes}
\end{figure}

\begin{figure}[h]
    \centering
    \includegraphics[scale=0.25]{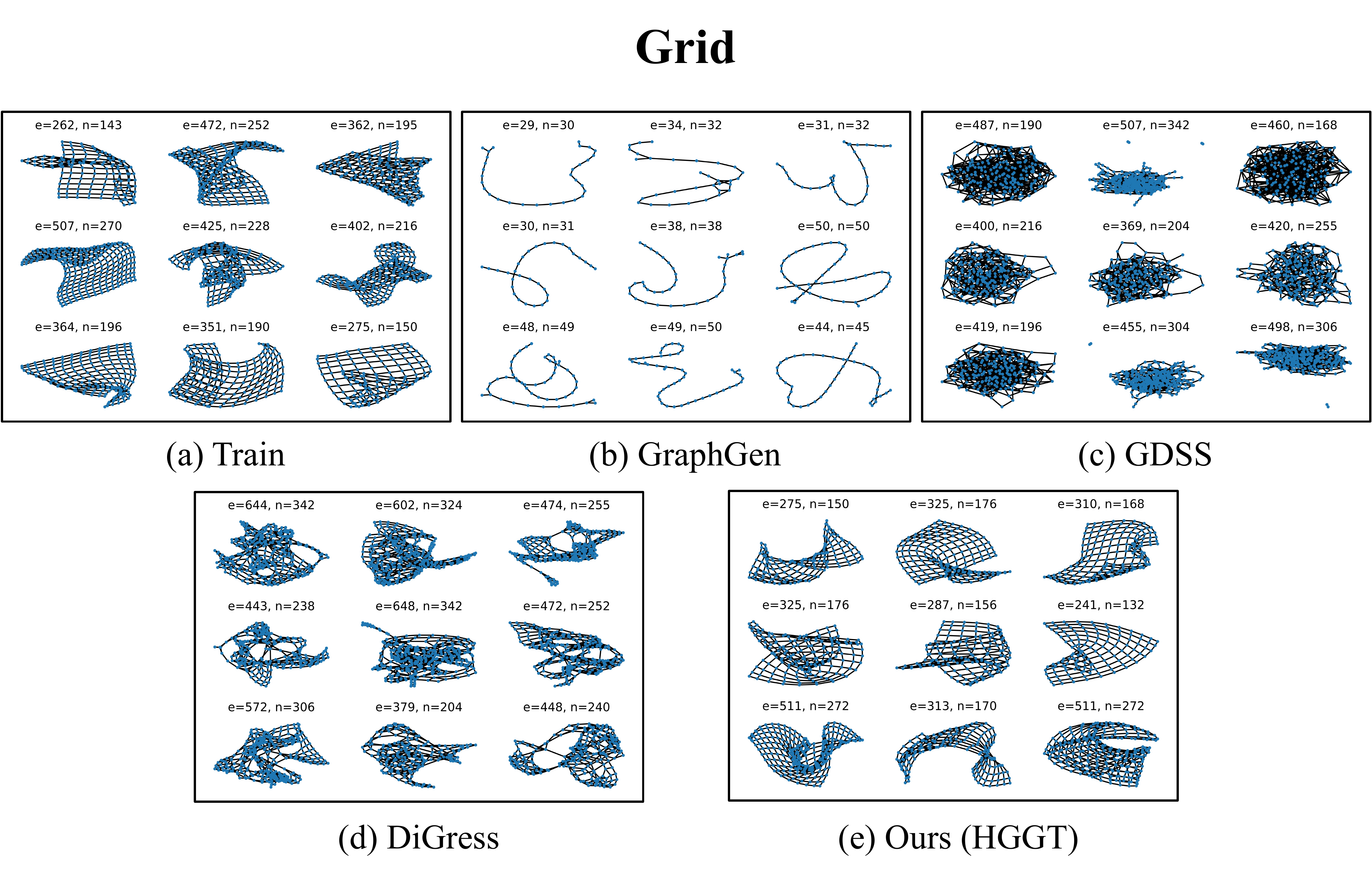}
    \vspace{-0.1in}
    \caption{\small \textbf{Visualization of the graphs from the Grid dataset and the generated graphs.}}
    \label{fig:grid}
\end{figure}
We present visualizations of graphs from the training dataset and generated samples from GraphGen, DiGress, GDSS, and \Algname in \cref{fig:community_small}, \cref{fig:planar}, \cref{fig:enzymes}, and \cref{fig:grid}. Note that we reproduced GraphGen and DiGress using open-source codes while utilizing the provided checkpoints for GDSS. However, given that the checkpoints provided for GDSS do not include the Planar dataset, we have omitted GDSS samples for this dataset. We additionally give the number of nodes and edges of each graph.

\clearpage
\subsection{Molecular graph generation}
\begin{figure}[ht]
    \centering
    \includegraphics[width=\linewidth]{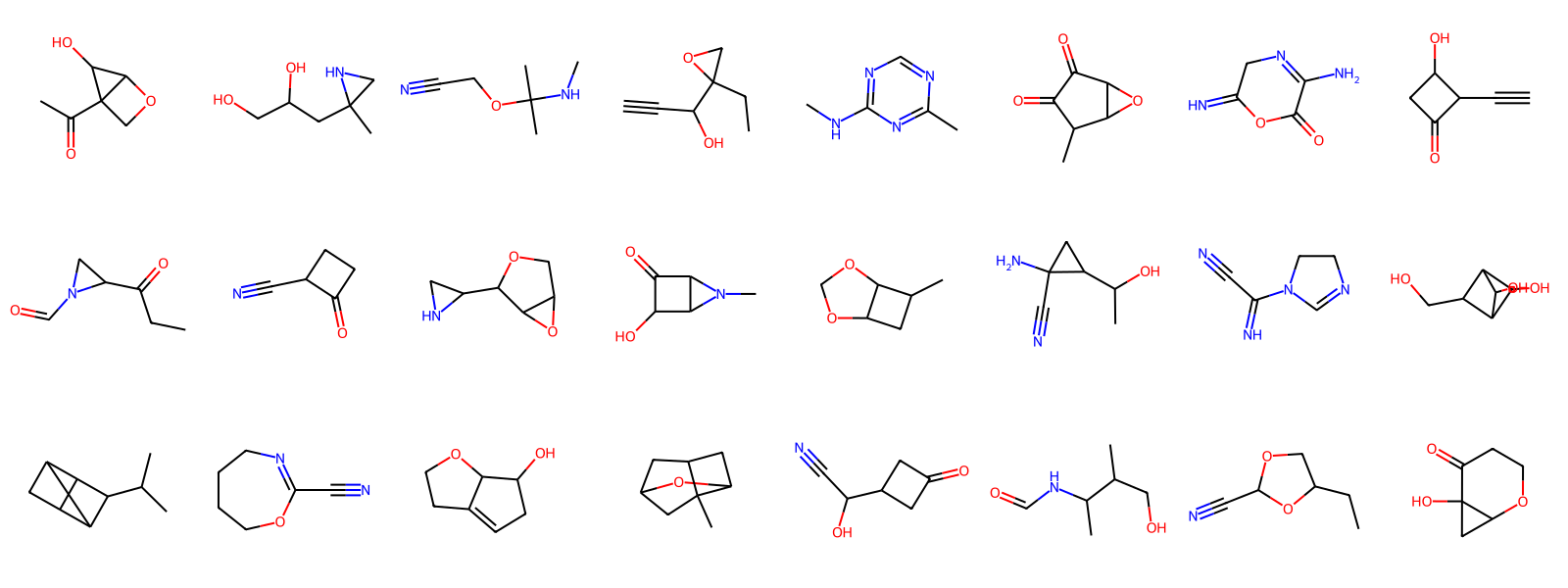}
    \vspace{-0.1in}
    \caption{\small \textbf{Visualization of the molecules generated from the QM9 dataset.}}
    \label{fig:qm9}
\end{figure}

\begin{figure}[ht]
    \centering
    \includegraphics[width=\linewidth]{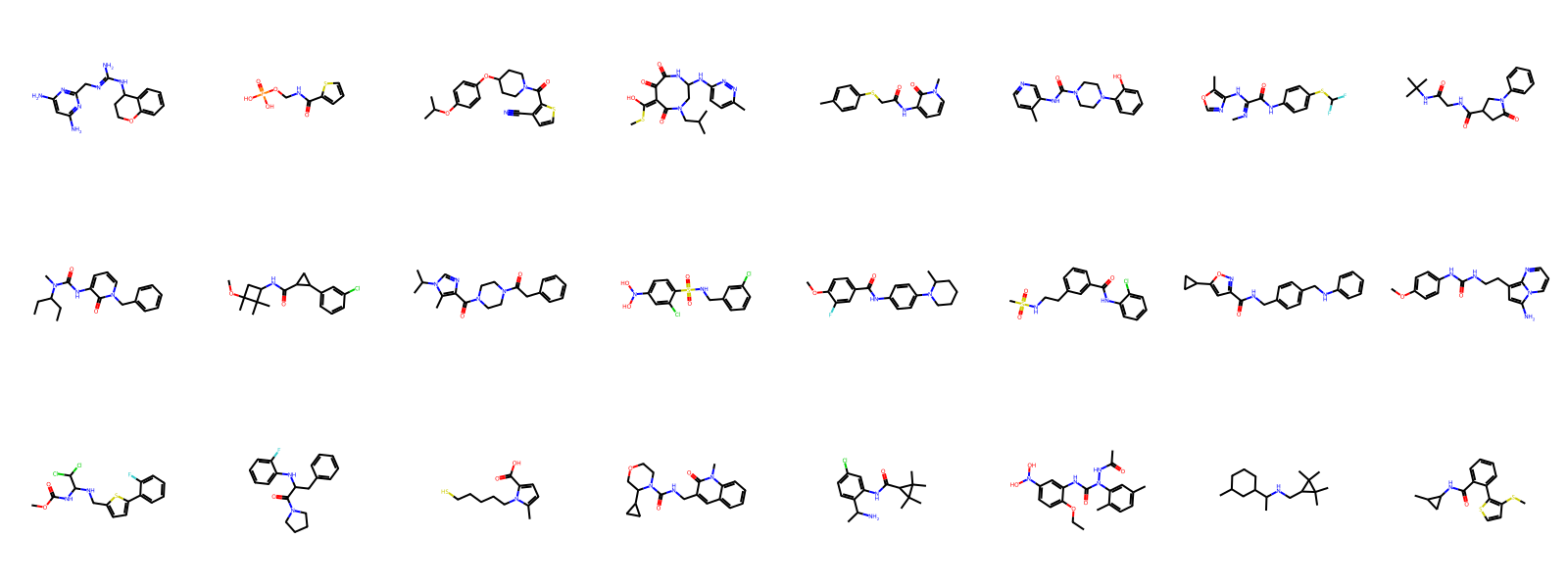}
    \vspace{-0.1in}
    \caption{\small \textbf{Visualization of the molecules generated from the ZINC250k dataset.}}
    \label{fig:zinc}
\end{figure}
We present visualizations of generated molecules from \Algname in \cref{fig:qm9} and \cref{fig:zinc}. Note that the 24 molecules are non-cherry-picked and randomly sampled.

\newpage
\section{Additional Experimental Results}\label{appx:additional_result}
In this section, we report additional experimental results.

%\begin{wraptable}{r}{0.55\textwidth}
\begin{figure}[h]
  \caption{\textbf{Generation results of \Algname with $k=3$}.}
  \label{tab:appx_additional}
  \centering
  \scalebox{0.8}{
    \begin{tabular}{ccccccc}
    \toprule
    & \multicolumn{3}{c}{{Community-small}} & \multicolumn{3}{c}{{Planar}} \\
    \cmidrule(lr){2-4} \cmidrule(lr){5-7}
    & Degree & Cluster. & Orbit & Degree & Cluster. & Orbit \\
    \midrule
    $k=2$ & \textbf{0.001} & \textbf{0.006} & 0.003 & \textbf{0.000} & \textbf{0.001} & \textbf{0.000} \\
    $k=3$ & 0.007 & 0.050 & \textbf{0.001} & 0.001 & 0.003 & \textbf{0.000} \\
    \bottomrule
  \end{tabular}
  }
\end{figure}
%\end{wraptable}

\subsection{Generic graph generation}
We provide generic graph generation results for $k=3$. Increasing $k$ decreases the sequence length, while vocabulary size increases to $2^{3^2}+2^6=578$.

We used Community-small and Planar datasets and measured MMD between the test graphs and generated graphs. We perform the same hyperparameter search for a fair evaluation as $k=2$. The results are in \cref{tab:appx_additional}. We can observe that \Algname still outperforms the baselines even with different $k$. 

\subsection{Molecular graph generation}\label{tab:xx}
% other metrics
% frag, intdiv, qed, sa, snn, scaffold, weight
\begin{table}[h]
  \caption{\textbf{Additional molecular graph generation performance.}}
  \label{tab:appx_mol}
  \centering
  \scalebox{0.9}{
    \begin{tabular}{lccccccc}
    \toprule & \multicolumn{7}{c}{ QM9 } \\ 
    \cmidrule(lr){2-8}
    Method & Frag. $\uparrow$ & Intdiv. $\uparrow$ & QED $\downarrow$ & SA $\downarrow$ & SNN $\uparrow$ & Scaf. $\uparrow$  & Weight $\downarrow$ \\
    \midrule
    DiGress & 0.9737 & \textbf{0.9189} & 0.0015 & \textbf{0.0189} & \textbf{0.5216} & 0.9063 & \textbf{0.1746} \\
    \midrule
    \Algname (Ours) & \textbf{0.9874} & 0.9150 & \textbf{0.0012} & 0.0304 & 0.5156 & \textbf{0.9368} & 0.2430 \\
    \bottomrule
    \end{tabular}
    }
    \vspace{-.15in}
\end{table}
\begin{table}[h]
    % \caption{\textbf{Additional molecular graph generation performance.}}
    %\label{tab:appx_mol}
    \centering
    \scalebox{0.9}{
    \begin{tabular}{lccccccc}
    \toprule & \multicolumn{7}{c}{ ZINC250k } \\ 
    \cmidrule(lr){2-8}
    Method & Frag. $\uparrow$ & Intdiv. $\uparrow$ & QED $\downarrow$ & SA $\downarrow$ & SNN $\uparrow$ & Scaf. $\uparrow$  & Weight $\downarrow$ \\
    \midrule
    DiGress & 0.7702 & \textbf{0.9061} & 0.1284 & 1.9290 & 0.2491 & 0.0001 & 62.9923 \\
    \midrule
    \Algname (Ours) & \textbf{0.9877} & 0.8644 &\textbf{0.0164} & \textbf{0.2407} & \textbf{0.4383} & \textbf{0.5298} & \textbf{1.8592} \\
    \bottomrule
    \end{tabular}
    }
\end{table}
We additionally report seven metrics of the generated molecules: (a) fragment similarity (Frag.), which measures the BRICS fragment frequency similarity between generated molecules and test molecules, (b) internal diversity (Intdiv.), which measures the chemical diversity in generated molecules, (c) quantitative estimation of drug-likeness (QED), which measures the drug-likeness similarity between generated molecules and test molecules, (d) synthetic accessibility score (SA), which compares the synthetic accessibility between generated molecules and test molecules, (e) similarity to the nearest neighbor (SNN), an average of Tanimoto similarity between the fingerprint of a generated molecule and test molecule, (f) scaffold similarity (Scaf.), the Bemis-Murcko scaffold frequency similarity between generated molecules and test molecules, and (g) weight, the atom weight similarity between generated molecules and test molecules. The results are in \cref{tab:appx_mol}.

\newpage

\textcolor{black}{\section{Discussion}\label{appx: discussion}}

\subsection{Hierarchy of \tree representation}\label{subsec: hierarchy}

\textcolor{black}{\tree representation is hierarchical as it forms a parent-child hierarchy between nodes. In detail, each node in \tree corresponds to a block (i.e., submatrix) in the adjacency matrix. Given a child and its parent node, the child node block is a submatrix of the parent block node, which enables \tree to represent a hierarchical structure between the blocks. While this hierarchy may differ from the exact hierarchical community structure, the \tree representation still represents a valid hierarchy present in the adjacency matrix.} \textcolor{black}{We also note that our \tree representation should not be confused with the hierarchical representation learned by graph neural networks with pooling functions \citep{ying2018hierarchical}.}

%existing term: hierarchical representation. The term ``hierarchical'' is used for the hierarchical community structure in prior works .  
%In detail, \cite{ying2018hierarchical} proposed to learn the hierarchical representation of graphs by learning soft cluster assignments of nodes and hierarchical pooling of node embeddings. We clarify that our \tree representation does not necessarily  capture the hierarchical community structure, while it captures an inherent hierarchy in the adjacency matrix.}

\textcolor{black}{Nevertheless, bandwidth minimization algorithms (including C-M node ordering) often induce node orderings that align with underlying clusters. Prior works \citep{barik2020vertex, mueller2004sparse} have empirically figured out that bandwidth minimization tends to cluster the points along the diagonal, which leads to a partial capture of the underlying community structure. This also supports our statement that \tree representation is hierarchical.} 

\subsection{Comparison with prior works on autoregressive graph generative model}

\textcolor{black}{In this section, we compare HGGT to two prior works on autoregressive graph generative models: GraphRNN \citep{you2018graphrnn} and GRAN \citep{liao2019efficient}. The main difference comes from the key idea of \Algname: the ability to capture recurring patterns and the hierarchy of the adjacency matrix. In detail, \Algname maps the recurring patterns in the dataset (large zero-filled block matrices) into a simple object (a zero-valued node in the \tree that the model can easily generate. This mapping allows the generative model to focus on learning instance-specific details rather than the generation of the whole pattern that is common across the dataset. In addition, \tree representation can represent a valid hierarchy present in the adjacency matrix, as described in \cref{subsec: hierarchy}}

\begin{figure}[h]
    \centering
\includegraphics[width=0.6\linewidth]{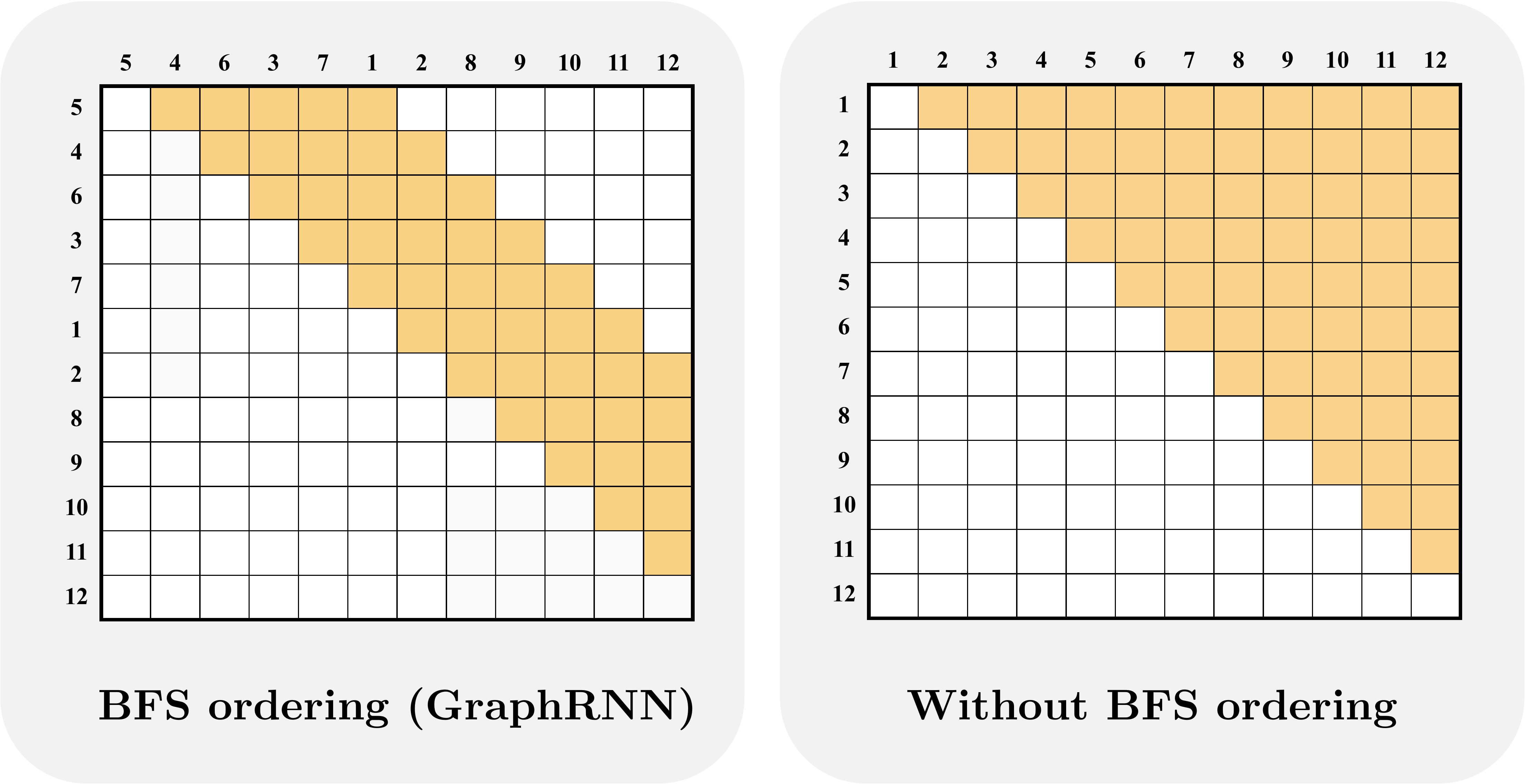}
    \vspace{-0.1in}
    \caption{\small \textbf{Reduced representation size of GraphRNN.}}
    \label{fig: graphrnn}
\end{figure}

\textcolor{black}{\textbf{Comparison to GraphRNN.} Both \Algname and GraphRNN reduce the representation size, which removes the burden of the graph generative models learning long-range dependencies. In detail, \Algname reduces the representation size by leveraging \tree, pruning, and tokenization. Otherwise, GraphRNN employs BFS node ordering constraining the upper-corner elements of the adjacency matrix to be consecutively zero as described in \cref{fig: graphrnn}.}

\begin{table}[h]
\centering
    \begin{tabular}{ccccc}
            \toprule 
             & Comm. & Planar & Enzymes & Grid \\
            \midrule
            Full matrix  & 241.3 & 4096.0 & 1301.6 & 50356.1 \\
            GraphRNN & \phantom{0}75.2 & 2007.3 & \phantom{0}234.7 & \phantom{0}4106.3 \\
            \Algname (ours) & \phantom{0}\textbf{30.3} & \phantom{0}\textbf{211.7} & \phantom{00}\textbf{67.3} & \phantom{00}\textbf{419.1} \\
            \bottomrule
        \end{tabular}
        \caption{\textbf{Representation size of GraphRNN and \Algname}}
        \label{tab: rep_size}
\end{table}

\textcolor{black}{The reduction of \Algname is higher as shown in \cref{tab: rep_size} that reports the average size of the representation empirically. Note that the representation size of \Algname and GraphRNN indicates the number of tokens and the number of elements limited by the maximum size of the BFS queue, respectively. While the comparison is not fair due to different vocabulary sizes, one could expect \Algname to suffer less from the long-range dependency problem due to the shorter representation.}

\textcolor{black}{\textbf{Comparison to GRAN.} The main difference between \Algname and GRAN comes from the different generated representations. \Algname generates \tree representation with large zero-filled blocks, which is further summarized into a single node in the \tree while GRAN generates the conventional adjacency matrix. Notably, the block of nodes of GRAN is solely used for parallel decoding, which is conceptually irrelevant to the graph representation.}

\textcolor{black}{In addition, the concept of block and the decoding process of the blocks differ in both methods. On one hand, the square-shaped \Algname block defines connectivity between a pair of equally-sized node sets. \Algname sequentially specifies these block matrix elements in a hierarchical way, i.e., first specifying whether the whole block matrix is filled with zeros and then specifying its smaller submatrices. On the other hand, the rectangular-shaped GRAN block defines connectivity between a set of newly added nodes and the existing nodes. GRAN can optionally decode the block matrix elements in parallel, speeding up the decoding process at the cost of lower performance.}

\begin{table}
    \centering
    \begin{tabular}{lccc}
    \toprule
      Dataset   & Rep. size & $N^2$  \\
      \midrule
       Comm.  & 48 & 400\\
       Enzymes & 238 & 15625\\
       Planar & 230 & 4696 \\
       Grid & 706 & 130321 \\
       \bottomrule
    \end{tabular}
    \label{tab:my_label}
\end{table}

\end{document}